\documentclass[a4paper,fleqn]{cas-dc}
\usepackage{amssymb}
\usepackage{amsmath}
\usepackage{graphicx}
\usepackage{booktabs}
\usepackage{longtable}
\usepackage{array}
\usepackage{enumitem}
\usepackage{tabularx}
\usepackage[authoryear]{natbib}
\newcolumntype{Y}{>{\raggedright\arraybackslash}X}
\newcolumntype{Z}[1]{>{\centering\arraybackslash}p{#1}}
\newcommand{\singlecoltablewidth}{0.49\textwidth}
\begin{document}
\let\WriteBookmarks\relax
\def\floatpagepagefraction{1}
\def\textpagefraction{.001}
\emergencystretch=2em
\sloppy

\shorttitle{RABC-Net for Low-Resource Dermoscopy}
\shortauthors{Y. Yao et~al.}
\RenewDocumentCommand \printorcid { } {}
\title[mode=title]{RABC-Net: Reliability-Aware Annotation-Free Skin Lesion Segmentation for Low-Resource Dermoscopy}

\author[1]{Yujie Yao}
\ead{202308437@stu.sicau.edu.cn}

\author[1]{Yuhaohang He}
\ead{heyuhaohang@stu.sicau.edu.cn}

\author[1]{Junjie Huang}
\ead{huangjunjie1@stu.sicau.edu.cn}

\author[1]{Zhou Liu}
\ead{liuzhou@stu.sicau.edu.cn}

\author[1]{Jiangzhao Li}
\ead{ljz030201@163.com}

\author[1]{Yan Qiao}
\ead{18481938015@163.com}

\author[1]{Wen Xiao}
\ead{202307322@stu.sicau.edu.cn}

\author[1]{Yunsen Liang}
\ead{202308549@stu.sicau.edu.cn}

\author[1]{Xiaofan Li}
\cormark[1]
\ead{xfl@sicau.edu.cn}

\affiliation[1]{organization={College of Information Engineering, Sichuan Agricultural University},
                city={Ya'an},
                country={China}}

\cortext[cor1]{Corresponding author}

\begin{abstract}
Pixel-level annotation is costly in low-resource dermoscopy. We present RABC-Net, a reliability-aware annotation-free segmentation system that combines pseudo-label reliability learning, restricted target-domain adaptation, and Reliability-Adaptive Boundary Calibration (RABC). The system decouples reliability learning from deployment: uncertainty-aware pseudo-label interaction shapes robust representations during training, while the image-only inference path is preserved and RABC performs local logit-space calibration from boundary confidence, uncertainty, and foreground probability. No manual masks are used for training or target-domain adaptation; validation labels, when available, are used only for final operating-point selection. Across ISIC-2017, ISIC-2018, and PH2, RABC-Net achieves macro-average DICE/JAC of 86.58\%/79.47\% and consistent matched-protocol results. Controlled within-study analyses show that RABC provides localized gains over nonlearned boundary correction, while the overall result comes from the full reliability-aware system. Adaptation updates only 3.50\% of model parameters, image-only inference runs at 87.4 FPS, and the selected operating points use $\sigma=0$ on all three datasets, indicating that learned calibration avoids extra smoothing at deployment.
\end{abstract}

\begin{keywords}
Skin lesion segmentation \sep annotation-free learning \sep boundary calibration \sep domain adaptation \sep low-resource deployment
\end{keywords}

\maketitle

\section{Introduction}

Automated skin lesion segmentation supports dermoscopic assessment of shape, color variation, and boundary irregularity. However, high-quality pixel-level masks remain expensive to scale across datasets, devices, and clinical sites.

Recent annotation-free and unsupervised methods improve lesion localization through clustering, saliency priors, self-supervised learning, and uncertainty-aware pseudo-label modeling \citep{ref17,ref19,ref22,ref23,ref20,ref24,ref25,ref1,ref26}. Three problems remain: pseudo-labels contain structured noise, boundary probabilities are often conservative, and pseudo-label reliability may shift under limited target-domain data.

Thus, better pseudo-label denoising does not automatically produce better contours. Fixed smoothing or binary dilation can improve recall, but these operations are not tied to local reliability and may introduce unnecessary foreground expansion. Deployable boundary correction should instead depend on local confidence and uncertainty.

We therefore propose RABC-Net, a reliability-aware annotation-free system for skin lesion segmentation. Its core module, Reliability-Adaptive Boundary Calibration (RABC), applies local logit-space correction using decoder features, boundary confidence, uncertainty, and foreground probability.

The contribution is the full reliability-aware deployment system rather than a single post-processing trick. RABC performs local calibration without changing the image-only inference path, while pseudo-label interaction, restricted adaptation, a stronger backbone, and feature-decoupled decoding provide stable reliability learning. Although pseudo-label initialization uses dermoscopy-specific priors, the broader framework centers on deployable reliability estimation, restricted adaptation, and local boundary correction. Experiments on ISIC-2017, ISIC-2018, and PH2 show consistent matched-protocol results, and controlled analyses isolate the narrower role of RABC within the system.

The main contributions of this work are summarized as follows:

\begin{enumerate}[leftmargin=*]
\item We present RABC-Net, a reliability-aware annotation-free deployment system that preserves an image-only inference path.
\item We combine reliability-aware pseudo-label learning, restricted interaction-branch adaptation, and lightweight RABC for local boundary calibration.
\item We report matched-protocol positioning and controlled within-study evidence on ISIC-2017, ISIC-2018, and PH2, including anchor comparisons, dilation controls, repeated-seed tests, and target-label-free evaluations.
\end{enumerate}

\section{Related Work}

\subsection{Unsupervised Skin Lesion Segmentation}

Early unsupervised dermoscopic segmentation relied mainly on low-level grouping cues. \citet{ref17} merged superpixels under hand-crafted similarity assumptions, while K-means \citep{ref23} and NCut \citep{ref22} represent the classical clustering family that partitions pixels or features without semantic supervision. Later methods injected stronger lesion priors. Saliency-CCE \citep{ref19} couples colour-context extraction with saliency reasoning, and texture-guided saliency distilling \citep{ref20} shows that richer appearance cues can sharpen unsupervised foreground localization. More recent approaches improve the feature space before mask extraction. \citet{ref24} learn spatially guided self-supervised clustering features, whereas \citet{ref25} impose lesion structure by minimizing structural entropy on multi-scale superpixel graphs. Classical refiners such as dense CRF \citep{ref40} and guided filtering \citep{ref41} are also widely used to sharpen coarse boundaries, but they are typically added as generic postprocessing rather than being tied to an explicit reliability model. Recent uncertainty-aware dermoscopic methods, notably RPI-Net \citep{ref1} and USL-Net \citep{ref26}, further improve pseudo-label quality, yet their emphasis remains denoising within a single domain. Our framework inherits reliability-aware pseudo-label interaction as one component, but extends the problem to cross-domain transfer and inference-stage boundary calibration.

\subsection{Cross-Domain Transfer and Uncertainty Modeling}

Cross-domain medical segmentation has been studied from several angles. \citet{ref35} adapt segmentation networks at test time to improve robustness without full retraining. \citet{ref36} align images and features to reduce modality gaps, while \citet{ref37} use shape-aware meta-learning to preserve anatomical structure under unseen-domain shift. In a stricter deployment setting, \citet{ref39} remove source data entirely and perform source-free adaptation from a pretrained model. Together, these studies show that domain shift can be mitigated, but many solutions still update a substantial part of the deployed segmentation network or introduce relatively heavy alignment machinery. In parallel, uncertainty modeling has become a standard tool for discounting noisy supervision and reweighting ambiguous pixels. Within dermoscopy, uncertainty-aware annotation-free learning has likewise been used to reweight unreliable pseudo-label paths and stabilize self-learning \citep{ref1,ref26}. Our setting lies at the intersection of these two lines of work, but differs in what is adapted: instead of updating the deployed image-only path, we freeze it and recalibrate only the pseudo-label interaction branch. This design reflects our hypothesis that the dominant cross-domain shift in annotation-free transfer lies in pseudo-label reliability topology rather than in the lesion semantics already captured by the source-trained backbone.

\subsection{Foundation Models and Lightweight Visual Backbones}

Foundation-model-based medical segmentation has expanded rapidly after SAM \citep{ref28}. MedSAM \citep{ref27} shows that promptable segmentation can transfer effectively to medical images after domain-specific adaptation, while SAMed \citep{ref31} and SAM-Med2D \citep{ref32} further tailor the SAM family to medical fine-tuning and broader 2D medical coverage. These models are important references, but they target a different operating regime from ours because prompts, extra supervision, or heavier fine-tuning are usually part of the workflow. We therefore treat them as reference methods from a different operating regime rather than direct annotation-free counterparts. On the backbone side, ConvNeXt \citep{ref2} shows that modern ConvNet design can provide stronger hierarchical features than legacy CNN encoders. This matters here because weak representations amplify pseudo-label noise and limit lesion-boundary recovery. Nevertheless, a stronger encoder alone does not resolve the coupled problems of noisy pseudo-label supervision, cross-domain reliability shift, and boundary probability compression. We therefore use ConvNeXt-Tiny as a stronger image backbone, while placing particular methodological emphasis on RABC and treating pseudo-label interaction together with parameter-decoupled adaptation as supporting mechanisms.

\section{Method}

\subsection{Overall Architecture Overview}

\begin{figure*}[t]
\centering
\includegraphics[width=\textwidth]{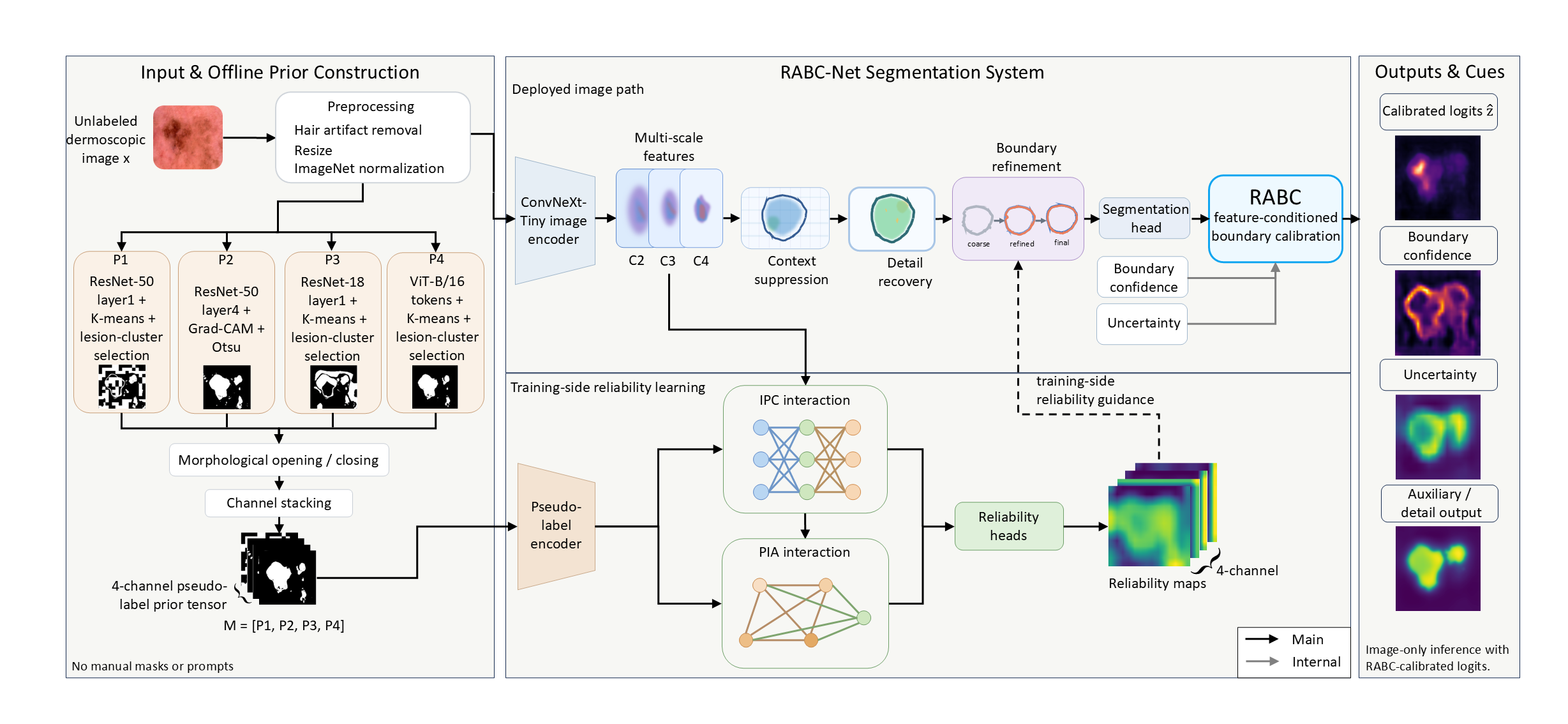}
\caption{Overall method overview. Four unsupervised paths produce a 4-channel pseudo-label prior tensor for IPC/PIA interaction. The image branch extracts multi-scale features, reliability heads estimate pseudo-label reliability, and the deployed image path uses RABC to calibrate logits before thresholding.}
\label{fig:method-overview}
\end{figure*}

\begin{figure*}[t]
\centering
\includegraphics[width=0.92\textwidth]{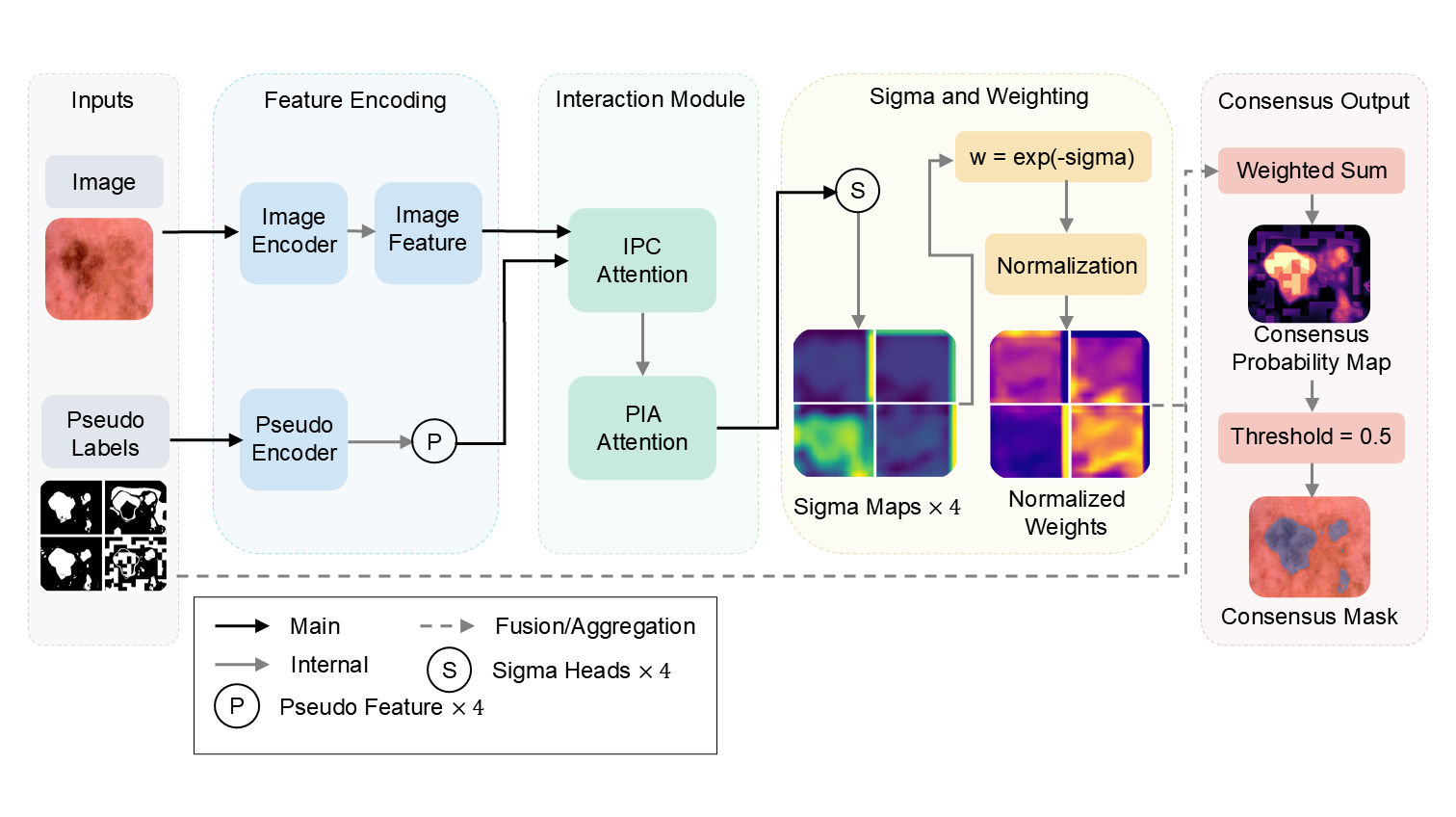}
\caption{Uncertainty-guided IPC/PIA correction. Image features and pseudo-label features interact through IPC and PIA attention, sigma heads produce reliability-aware weights, and the resulting weighted consensus probability map is binarized into a cleaner consensus mask for supervision.}
\label{fig:ipcpia-correction}
\end{figure*}

\begin{figure*}[t]
\centering
\includegraphics[width=0.96\textwidth]{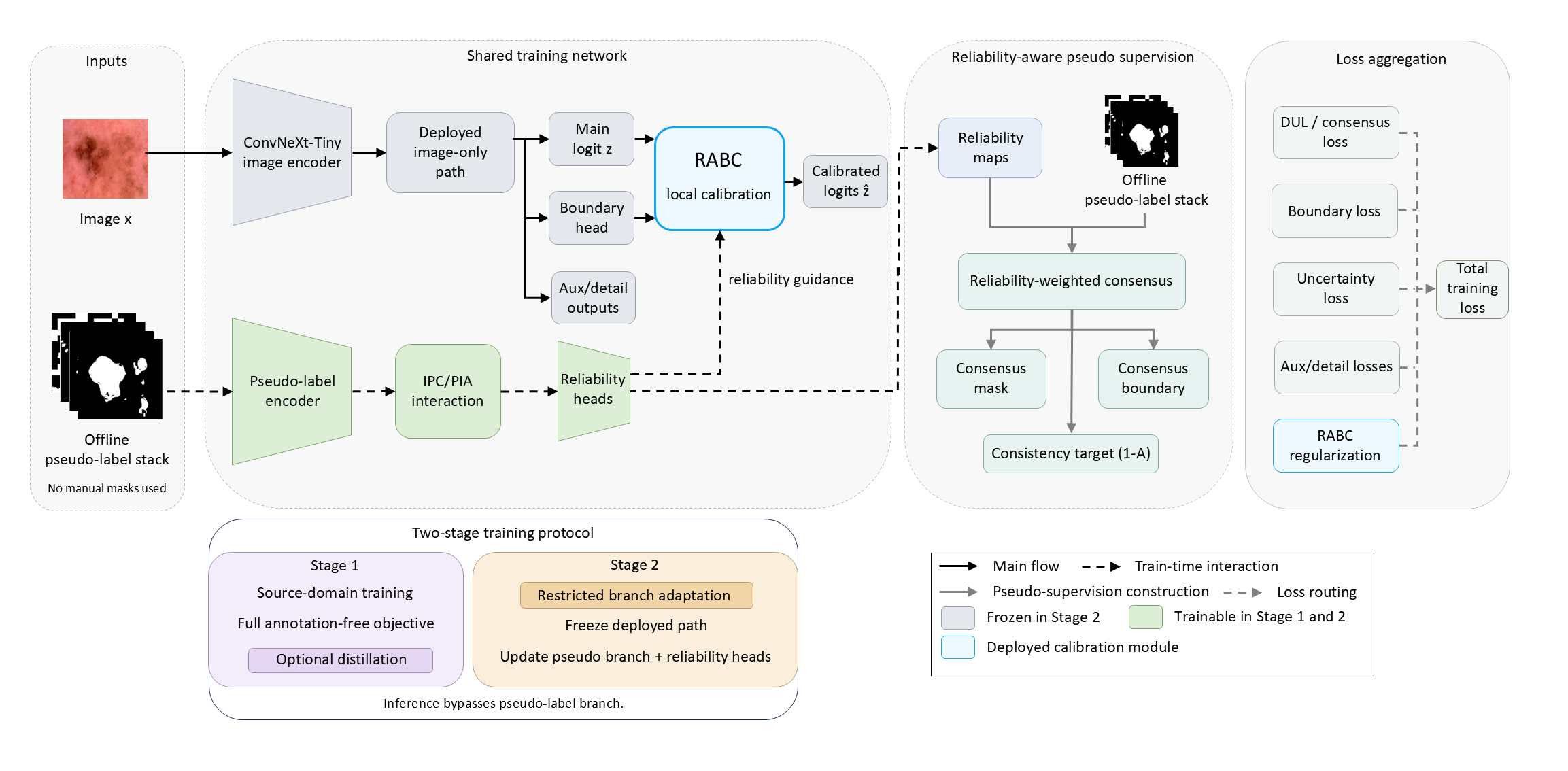}
\caption{Training-stage view of RABC-Net. The figure shows the shared forward path, reliability-weighted pseudo supervision, restricted branch adaptation, and the loss groups used for annotation-free optimization.}
\label{fig:training-strategy}
\end{figure*}

\begin{figure*}[t]
\centering
\includegraphics[width=0.96\textwidth]{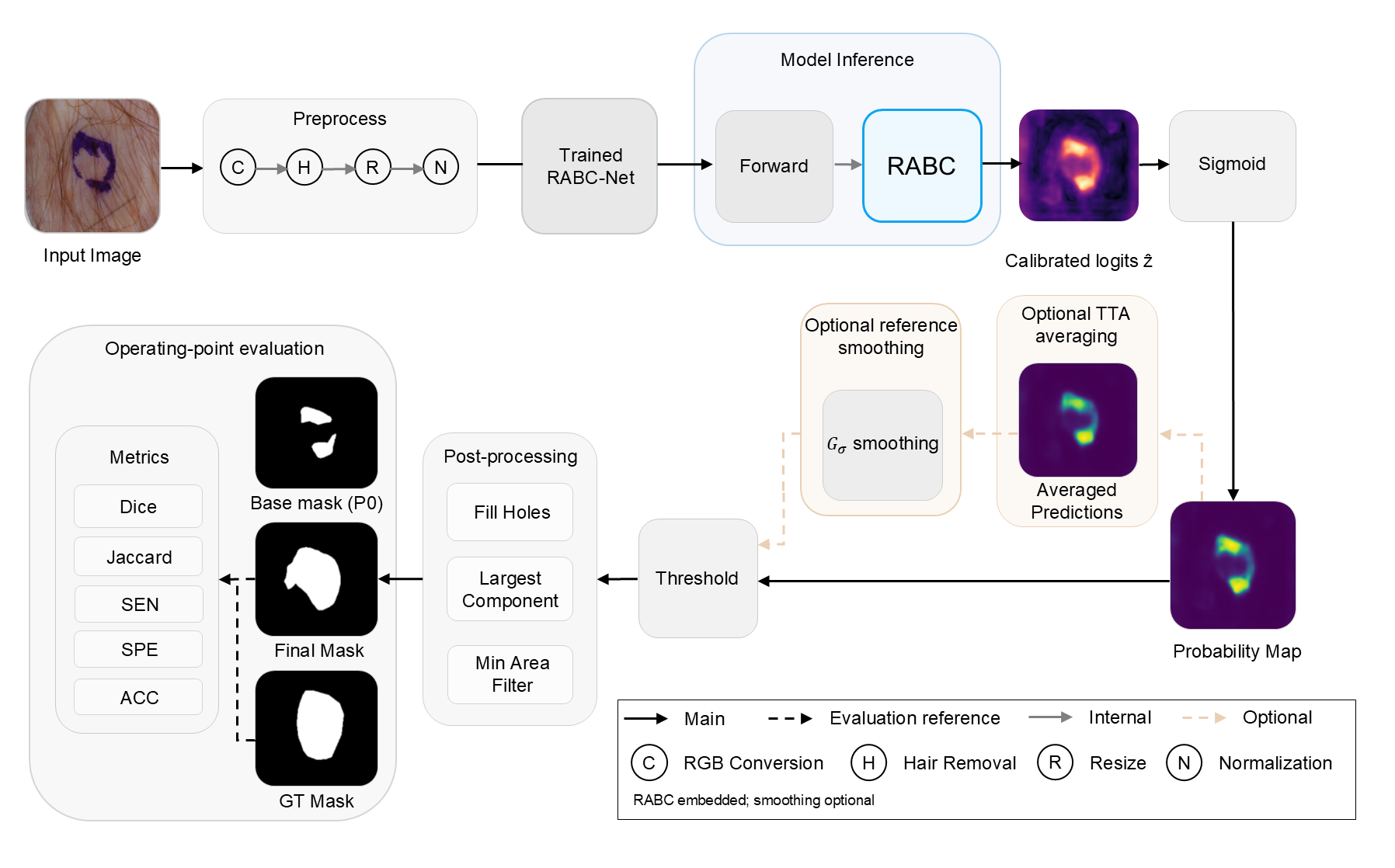}
\caption{Inference-stage operating-point pipeline. RABC-calibrated logits are converted to a probability map, optional TTA and reference smoothing are applied only when selected, and lightweight morphology yields the final mask.}
\label{fig:gps-pipeline}
\end{figure*}

RABC-Net combines training-side reliability learning with inference-side boundary calibration. Pseudo-label interaction and uncertainty modeling stabilize annotation-free training, while RABC predicts local logit corrections from decoder features, boundary confidence, uncertainty, and foreground probability. Figure~\ref{fig:method-overview} summarizes the workflow; Figures~\ref{fig:ipcpia-correction}, \ref{fig:training-strategy}, and \ref{fig:gps-pipeline} detail the interaction, training, and inference stages.

The system extends prior reliability-aware annotation-free segmentation with interaction-branch adaptation, a stronger image backbone/decoder, and lightweight RABC calibration.

\subsection{Problem Definition}

Given an unlabeled image set $\mathcal{X}=\{x_n\}_{n=1}^{N}$, the goal is to learn a segmentation model $f_\theta$ that predicts a lesion probability map $\hat{y}=f_\theta(x)\in[0,1]^{H\times W}$. No ground-truth labels are used at any point during training; all supervision is derived from automatically generated multi-source pseudo-labels and auxiliary self-supervised objectives.

\subsection{Multi-Source Pseudo-Labels and Consistency Mapping}

For each image, $K=4$ pseudo-label paths are constructed from features extracted by different pretrained backbones, each providing a complementary segmentation prior (Table~\ref{tab:pseudo-label-paths}):

\begin{table*}[t]
\centering
\scriptsize
\setlength{\tabcolsep}{2pt}
\renewcommand{\arraystretch}{1.08}
\caption{Pseudo-label path definitions and feature sources.}\label{tab:pseudo-label-paths}
\begin{tabularx}{\textwidth}{@{}Z{0.10\textwidth}YY@{}}
\toprule
Path & Feature Source & Binarization Method \\
\midrule
$P_1$ & ResNet-50 \texttt{layer1} spatial features & K-means ($k=2$) + lesion cluster selection \\
$P_2$ & ResNet-50 \texttt{layer4} Grad-CAM heatmap & Otsu thresholding \\
$P_3$ & ResNet-18 \texttt{layer1} spatial features & K-means ($k=2$) + lesion cluster selection \\
$P_4$ & ViT-B/16 patch token features & K-means ($k=2$) + lesion cluster selection \\
\bottomrule
\end{tabularx}
\end{table*}

Paths $P_1$/$P_3$/$P_4$ promote diversity by applying K-means clustering to feature maps of architecturally distinct networks (CNN vs. Transformer) at different representational depths; $P_2$ provides a complementary gradient-based perspective via Grad-CAM activation maps followed by Otsu thresholding. In the present dermoscopy instantiation, foreground-cluster selection uses fixed brightness, centrality, and area heuristics only to choose among unsupervised proposals. The transferable core of the framework lies in the subsequent reliability learning, restricted adaptation, and local calibration stages, for which alternative prior generators could be substituted in other imaging settings. No SAM-based or prompt-based path is used in the main training pipeline. All masks are refined with morphological opening and closing operations, then stacked as in Eq.~\ref{eq:pseudo-stack}.
\begin{equation}
\mathbf{P}=\{P_i\}_{i=1}^{K},\quad P_i\in[0,1]^{H\times W}
\label{eq:pseudo-stack}
\end{equation}
The pseudo-label consensus map is defined in Eq.~\ref{eq:pseudo-consensus}.
\begin{equation}
P_c = \frac{1}{K}\sum_{i=1}^{K}P_i
\label{eq:pseudo-consensus}
\end{equation}
Entropy-based pixel consistency is defined in Eq.~\ref{eq:pixel-consistency}.
\begin{equation}
A = 1 - \frac{-P_c\log(P_c+\epsilon)-(1-P_c)\log(1-P_c+\epsilon)}{\ln 2}
\label{eq:pixel-consistency}
\end{equation}
where $A\in[0,1]$; higher values indicate greater inter-source agreement.

\subsection{IPC/PIA Uncertainty Interaction Mechanism}

To provide a reliable feature space and structural basis for the later boundary probability calibration stage, we adopt the two-stage pseudo-label feature interaction mechanism of RPI-Net \citep{ref1} to exploit the complementary information in multi-source pseudo-labels and to explicitly model their reliability differences:

\begin{itemize}[leftmargin=*]
\item \textbf{IPC (Image--Pseudo Cross-attention)}: Image features serve as queries, and each pseudo-label feature map serves as keys/values; scaled dot-product attention fuses image semantics with pseudo-label priors across modalities.
\item \textbf{PIA (Pseudo--Pseudo Interaction)}: Pseudo-label features from different sources interact via cross-attention to enhance consistent regions and resolve contradictions in ambiguous regions.
\end{itemize}

After each interaction stage, the fused features pass through an independent $\sigma_i$ convolutional head that predicts a pixel-level uncertainty map, used to compute dynamic loss weights in DUL. Our framework extends this design in two ways: (1) during the Stage 2 interaction-branch adaptation protocol (Section 3.8), the pseudo-label encoder and IPC/PIA modules remain trainable, allowing them to adapt to the target domain's pseudo-label noise distribution during training; (2) paired with the stronger ConvNeXt-Tiny backbone, the interaction attention can capture finer-grained semantic correspondences.

\subsection{Dynamic Uncertainty Loss (DUL)}

We adopt the Dynamic Uncertainty Loss (DUL) of RPI-Net \citep{ref1}. The pixel-level loss for the $i$-th pseudo-label path is given in Eq.~\ref{eq:dul-path-loss}.
\begin{equation}
\mathcal{L}_i = \mathbb{E}_{b,h,w}\left[\frac{\mathrm{BCE_{logits}}(z,P_i)}{\exp(\sigma_i)} + \frac{1}{2}\sigma_i\right]
\label{eq:dul-path-loss}
\end{equation}
where $z\in\mathbb{R}^{B\times 1\times H\times W}$ denotes the segmentation logit (pre-sigmoid), $\mathrm{BCE_{logits}}$ is the binary cross-entropy with built-in sigmoid, $\sigma_i\in\mathbb{R}^{B\times 1\times H\times W}$ is the pixel-level log-variance map predicted by an independent convolutional head for path $i$, and $\mathbb{E}_{b,h,w}$ is the mean over batch and spatial dimensions.

When consistency weighting is enabled, we set $\gamma=1$ in all reported experiments, so the BCE term is elementwise-multiplied by $A$. The path-level weight is given in Eq.~\ref{eq:dul-path-weight}.
\begin{equation}
\alpha_i = \frac{1}{\mathbb{E}[(\mathrm{softplus}(\sigma_i))^2]+\epsilon}
\label{eq:dul-path-weight}
\end{equation}
The aggregated DUL objective is given in Eq.~\ref{eq:dul-total}.
\begin{equation}
\mathcal{L}_{dul}=\frac{\sum_{i=1}^{K}\alpha_i\mathcal{L}_i}{\sum_{i=1}^{K}\alpha_i+\epsilon}
\label{eq:dul-total}
\end{equation}

\subsection{Other Training Losses}

The total training loss is a weighted combination of the following terms:

\begin{itemize}[leftmargin=*]
\item Pseudo-label consensus segmentation loss (BCE + Dice + optional Tversky).
\item Boundary alignment loss (L1 between the predicted boundary map and the Sobel-derived pseudo-label consensus boundary).
\item Boundary head supervision loss (BCE).
\item Uncertainty head supervision loss (BCE, targeting $1-A$).
\item Optional auxiliary head and detail head consistency losses.
\end{itemize}

The annotation-free training constraint is enforced by setting $w_{\text{sup}}=0$ and labeled sample ratio $r_{\text{labeled}}=0$ throughout training. The exact weighting coefficients for these loss terms during source-domain training and target-domain adaptation are detailed in Section 4.2.

\subsection{Source Domain Distillation Training}

To transfer segmentation priors from the source dataset ISIC-2017 to target domains, we train a frozen teacher $f_{\theta^*}$ on ISIC-2017 and use its soft predictions as auxiliary supervision on target images, as defined in Eq.~\ref{eq:distill-loss}.
\begin{equation}
\mathcal{L}_{\text{distill}} = \text{BCE}\big(z,\; \mathrm{Sigmoid}(f_{\theta^*}(x))\big)
\label{eq:distill-loss}
\end{equation}
The distillation weight follows the Gaussian ramp-up schedule in Eq.~\ref{eq:distill-weight}.
\begin{equation}
\begin{aligned}
w_{\text{distill}}(t)
&= w_{\text{distill}}^{\max}\exp\!\left(-5(1-r(t))^2\right), \\
r(t) &= \min\!\left(1, \frac{t+1}{T_{\text{rampup}}}\right)
\end{aligned}
\label{eq:distill-weight}
\end{equation}
so reliance on the teacher increases as the student stabilizes. This transfers source-domain knowledge in output space without feature alignment and yields $\Delta$DICE=+5.78 pp in the A0$\rightarrow$A1 ablation.

\subsection{Interaction-Branch Adaptation}

Stage 2 addresses cross-domain uncertainty shift by partitioning the model into a frozen deployed image-only segmentation path and a trainable pseudo-label interaction branch. The frozen set contains the ConvNeXt-Tiny encoder, decoder/refinement modules, segmentation head, and auxiliary boundary/detail/uncertainty heads. The trainable set contains the pseudo-label encoder, IPC/PIA modules, and $\sigma$ heads; when RABC is enabled, the compact calibration head is added to this set as well. In our ISIC-2018 RABC configuration, the trainable set contains 1.102 M of 31.468 M parameters (3.50\%). The rationale is that annotation-free cross-domain degradation is driven mainly by changes in pseudo-label reliability and disagreement structure, not by a collapse of generic lesion semantics in the image backbone. Fully updating the deployed image-only path on a small target set would therefore risk imprinting target-domain pseudo-label errors onto the mapping used at test time. By restricting adaptation to pseudo-label encoding, interaction, uncertainty weighting, and calibration, Stage 2 changes how noisy target-domain supervision is interpreted during optimization while preserving the source-trained image-to-mask function used at inference. The gradient restriction for the trainable set $\mathcal{T}$ is given in Eq.~\ref{eq:stage2-gradient}.
\begin{equation}
\nabla_{\theta_m} = \begin{cases} \nabla_{\theta_m}\mathcal{L} & \text{if } m \in \mathcal{T} \\ 0 & \text{otherwise} \end{cases}
\label{eq:stage2-gradient}
\end{equation}
BatchNorm layers outside the trainable-prefix set remain in evaluation mode. Stage 2 optimizes the same unsupervised objective as Stage 1 but restricts gradients to $\mathcal{T}$. Because test-time evaluation uses \texttt{model(x, None)}, the pseudo-label interaction branch is bypassed at inference; we therefore interpret Stage 2 as training-side reliability recalibration rather than direct fine-tuning of the deployed image-only path. In the later training-side ablation, A2 should accordingly be read as the full Stage 2 protocol together with its validation-selected operating point.

\subsection{Feature-Decoupled Decoding Paradigm for Skin Lesion Boundaries}

The decoder stabilizes lesion-mask learning before boundary calibration by separating context suppression, detail recovery, and boundary refinement. Channel-spatial attention suppresses background responses, a guided skip path restores low-contrast detail, and a boundary-focused block refines uncertain contours. The boundary branch uses the uncertainty-gated refinement in Eq.~\ref{eq:boundary-refine}.
\begin{equation}
\hat{z} = z + \mathcal{R}(\phi_s \cdot g),\quad g = \sigma\!\left(W_g \cdot [\phi_s;\, b;\, u]\right)
\label{eq:boundary-refine}
\end{equation}
where $b$ and $u$ are boundary and uncertainty cues from the main segmentation path. The complete decoder adds only 2.83 M parameters (9.9\% of the backbone), yet the leave-one-out decoder ablation reported later shows that all three module groups contribute materially to performance.

\subsection{Reliability-Adaptive Boundary Calibration (RABC)}

Conservative pseudo-supervision often weakens responses near uncertain contours. Lowering the global threshold can recover recall, but shifts all pixel decisions and may increase false positives. We instead introduce \textbf{Reliability-Adaptive Boundary Calibration (RABC)} as a learned local logit-space corrector.

Let $z$ denote the current segmentation logits, $p=\mathrm{Sigmoid}(z)$ the corresponding foreground probability, $b=\mathrm{Sigmoid}(z_b)$ the boundary confidence map, and $u=\mathrm{Sigmoid}(z_u)$ the uncertainty map. RABC first constructs a calibration candidate map that highlights uncertain boundary-adjacent pixels which are still predicted conservatively:
\begin{equation}
c = b \cdot (0.35 + 0.65u) \cdot (1-p)
\label{eq:rabc-candidate}
\end{equation}
The term $(1-p)$ suppresses already confident foreground pixels, while $b$ and $u$ focus the calibration on ambiguous contour neighborhoods.

Given decoder feature map $\phi$, a lightweight context head predicts a reliability-adaptive strength map $\alpha$, a threshold-shift term $\Delta\tau$, and a suppression gate $s$ from the concatenated cues $[\phi; b; u; p]$. A local mixer $\mathcal{M}(\cdot)$ averages nearby logits, and a residual head $r(\phi)$ provides feature-conditioned fine correction. The final RABC update is
\begin{equation}
\Delta z = \alpha \cdot c \cdot (\mathcal{M}(z)-z) + \Delta\tau - s \cdot \beta \cdot (1-b)(1-u)p + c \cdot r(\phi)
\label{eq:rabc-delta}
\end{equation}
\begin{equation}
\hat{z} = z + \Delta z
\label{eq:rabc-update}
\end{equation}
where $\beta$ is a learnable background-preservation scale. The first two terms encourage boundary recovery in boundary-adjacent, uncertain, and still under-activated locations. The third term suppresses far-background drift by down-weighting confident background regions. The residual term allows small feature-driven corrections when the hand-designed cues are insufficient.

In implementation terms, the context head is a compact $1{\times}1 \rightarrow 3{\times}3 \rightarrow 1{\times}1$ convolutional stack that outputs three single-channel pixel maps: $\alpha$, $\Delta\tau$, and $s$. The local mixer is a single $3{\times}3$ averaging convolution on logits, and the residual branch is a single $1{\times}1$ convolution on decoder features. For the reported ConvNeXt-Tiny setting with hidden dimension 128, the entire RABC module adds only 45{,}839 parameters ($\approx 0.046$M), so its role is to provide a lightweight learned calibration path rather than a second heavy decoder.

Compared with threshold reduction or binary dilation, RABC remains coupled to local probability structure. Dilation acts in binary-mask space with a fixed structuring element after thresholding, whereas RABC acts in logit space and predicts whether, where, and how strongly a correction should be applied. To keep these updates local, we regularize RABC through boundary-consistency, far-background-preservation, and sparsity losses. With pseudo-consensus map $P_c$, Sobel boundary magnitude $B=\mathcal{S}(P_c)$, boundary band $\bar{B}=\mathrm{MaxPool}(B,k=5)$, support map $\alpha$, and boundary weight $W_b=1+\lambda_b P_c\max(B,b)(0.5+0.5u)(1-p)$, we use
\begin{equation}
\mathcal{L}_{\text{rabc-bnd}}
= \mathrm{BCE}_{\text{logits}}\!\left(\hat{z},P_c; W_b\right)
\label{eq:rabc-bnd}
\end{equation}
\begin{equation}
\begin{aligned}
\mathcal{L}_{\text{rabc-far}}
= {} &
\frac{
\sum \mathrm{ReLU}\!\left(\sigma(\hat{z})-\sigma(z)-m\right)\,(1-P_c)(1-\bar{B})
}{
\sum (1-P_c)(1-\bar{B})+\epsilon
}
\end{aligned}
\label{eq:rabc-far}
\end{equation}
\begin{equation}
\mathcal{L}_{\text{rabc-sp}}
=
\mathrm{mean}\!\left(|\Delta z|\,(1-\alpha)\right)
\label{eq:rabc-sp}
\end{equation}
where $m=0.02$ and $k=5$ in all reported RABC experiments. The RABC-specific loss weights are set to 0.04, 0.02, and 0.01 for $\mathcal{L}_{\text{rabc-bnd}}$, $\mathcal{L}_{\text{rabc-far}}$, and $\mathcal{L}_{\text{rabc-sp}}$, respectively. In this sense, RABC is a reliability-aware calibration mechanism embedded within the segmentation network rather than a fixed post-threshold heuristic.

At inference, the deployed image branch directly outputs calibrated logits $\hat{z}$, which are converted to probabilities and thresholded in the standard way. Optional Gaussian smoothing is retained only as a lightweight reference comparator in the operating-point study, allowing us to test whether learned calibration already absorbs part of the effect that simple post-processing would otherwise provide.

\subsection{Inference Pipeline}

The complete inference pipeline comprises the following composable steps:

\begin{enumerate}[leftmargin=*]
\item \textbf{RABC prediction}: The image branch produces calibrated logits through the embedded RABC module and converts them to a probability map.
\item \textbf{TTA (Test-Time Augmentation)}: Apply $K$ flip transformations (horizontal/vertical) to the input image, forward pass each augmented view independently, and average the resulting probability maps. Modes: \texttt{none} ($K=1$), \texttt{flip2} ($K=2$, horizontal flip only), \texttt{flip4} ($K=4$, horizontal, vertical, and both combined).
\item \textbf{Optional reference smoothing}: Apply Gaussian smoothing $G_\sigma$ to the averaged probability map only when reference smoothing is enabled during operating-point selection (skipped when $\sigma=0$).
\item \textbf{Thresholding}: $\hat{m}=\mathbb{1}[\tilde{p}>\tau]$; the threshold $\tau$ is selected on the validation set based on the target evaluation metric.
\item \textbf{Morphological post-processing}: Optional \texttt{fill\_\allowbreak holes} and \texttt{keep\_\allowbreak largest} steps remove scattered noise and preserve the dominant connected component at inference time.
\end{enumerate}

The operating point ($\tau, \sigma$, TTA mode, and post-processing flags) for each dataset is determined entirely by grid search on the validation set; the test set does not participate in any selection or tuning. In this study, $\sigma$ is treated as an optional reference-smoothing flag rather than as the calibration mechanism itself.

\section{Experiments}

\subsection{Experimental Protocol}

We evaluate on three public dermoscopy datasets:

\begin{itemize}[leftmargin=*]
\item \textbf{ISIC-2017} \citep{ref3}: 2,000 training / 150 validation / 600 test images.
\item \textbf{ISIC-2018} \citep{ref4}: 2,594 training / 100 validation / 1,000 test images.
\item \textbf{PH2} \citep{ref5}: 200 images total; randomly split with seed=42 at ratios (0.70, 0.15, 0.15) into 140 training / 30 validation / 30 test images.
\end{itemize}

All experiments are implemented in PyTorch with AdamW ($\beta_1=0.9$, $\beta_2=0.999$, weight decay $10^{-4}$), gradient clipping at 1.0, and input resolution $224\times224$. Training augmentation is intentionally minimal---random horizontal and vertical flips only---because stronger photometric or geometric perturbations destabilize the contrast structure and spatial correspondence assumed by automatic pseudo-label generation. Source-domain training on ISIC-2017 is performed from scratch with learning rate $6\times10^{-6}$, batch size 16, 24 epochs, and one iterative pseudo-label refinement round (8 epochs, blend ratio 0.35). For target-domain adaptation on ISIC-2018 and PH2, the best Stage 1 checkpoint is reused, the deployed image-only path is frozen, and only the pseudo\_encoder, IPC/PIA modules, $\sigma$ heads, and the compact RABC head when enabled are updated. This second stage uses learning rate $5\times10^{-6}$, batch size 12, and 2 epochs without iterative refinement. Stage 1 source training uses pseudo-consensus/boundary/boundary-head/uncertainty/aux/detail weights of 0.15/0.03/0.15/0.08/0.05/0.08, while Stage 2 target adaptation uses 0.10/0.02/0.10/0.05/0.03/0.05. On ISIC-2018, the source-distillation stage uses weight 0.8 with a one-epoch Gaussian ramp-up and a frozen ISIC-2017 teacher checkpoint; the subsequent Stage 2 adaptation disables source distillation and reuses the best distilled checkpoint as initialization. Key differences between the two stages are summarized in Table~\ref{tab:train-stage-hparams}.

\begin{table*}[t]
\centering
\scriptsize
\setlength{\tabcolsep}{4pt}
\renewcommand{\arraystretch}{1.08}
\caption{Key differences between Stage 1 source training and Stage 2 interaction-branch adaptation.}\label{tab:train-stage-hparams}
\begin{tabular}{@{}>{\raggedright\arraybackslash}p{0.38\textwidth}@{\hspace{24pt}}>{\centering\arraybackslash}p{0.10\textwidth}@{\hspace{80pt}}>{\raggedright\arraybackslash}p{0.35\textwidth}@{}}
\toprule
Setting & Stage 1 & Stage 2 \\
\midrule
Training split & ISIC-2017 & ISIC-2018 / PH2 \\
Trainable modules & Full model & pseudo\_encoder + IPC/PIA + $\sigma$ heads \\
Frozen deployed image-only path & No & Yes \\
Learning rate & $6\times10^{-6}$ & $5\times10^{-6}$ \\
Batch size & 16 & 12 \\
Epochs & 24 & 2 \\
Iterative pseudo-label refinement & One round & None \\
\bottomrule
\end{tabular}
\end{table*}

Table~\ref{tab:model-complexity} reports model complexity and runtime statistics of the proposed framework.

\begin{table}[htbp]
\centering
\scriptsize
\setlength{\tabcolsep}{3pt}
\renewcommand{\arraystretch}{1.08}
\caption{Model complexity and runtime characteristics of the proposed framework. The highlighted rows summarize the main deployment takeaway: lightweight image-only inference, low memory use, and restricted adaptation overhead.}\label{tab:model-complexity}
\resizebox{\singlecoltablewidth}{!}{%
\begin{tabular}{@{}ll@{}}
\toprule
Metric & Value \\
\midrule
Total parameters & 31.47 M \\
ConvNeXt-Tiny backbone parameters & 28.59 M \\
Decoder/head additional parameters & 2.83 M (9.9\% of backbone) \\
GFLOPs (image inference only) & 57.39 \\
GFLOPs (with pseudo-label interaction) & 59.59 \\
GFLOPs (ConvNeXt-Tiny backbone only) & 8.91 \\
GPU inference speed (image only) & \textbf{87.4 FPS (11.4 ms/image)} \\
GPU inference speed (with pseudo-labels) & 55.3 FPS (18.1 ms/image) \\
Peak GPU memory & \textbf{686 MB} \\
\bottomrule
\end{tabular}
}
\end{table}

Measurements were obtained on an NVIDIA RTX 4060 Ti at input resolution $224\times224$. The RABC-enabled model contains 31.47 M parameters and the deployed image-only path remains lightweight in practice, corresponding to 29.9\% of the RPI-Net parameter count \citep{ref1}. During Stage 2 adaptation, only 1.102 M of 31.468 M parameters (3.50\%) are updated, so transfer remains restricted to the reliability side of the system rather than rewriting the deployed image-only mapping.

We report standard overlap metrics (ACC, DICE, JAC, SEN, SPE) and boundary distance metrics (HD95, ASSD). Metrics are computed per test image and then averaged over the test set. For HD95/ASSD, empty-empty pairs are assigned 0, whereas one-empty pairs receive a diagonal-length penalty.

Operating points are determined by validation-set grid search and held fixed for all test evaluations (Table~\ref{tab:operating-points}):

\begin{table}[htbp]
\centering
\scriptsize
\setlength{\tabcolsep}{3pt}
\renewcommand{\arraystretch}{1.08}
\caption{Validation-selected operating points used for final test evaluation.}\label{tab:operating-points}
\resizebox{\singlecoltablewidth}{!}{%
\begin{tabular}{@{}llcccc@{}}
\toprule
Dataset & $\tau$ & $\sigma$ & TTA & fill\_holes & keep\_largest \\
\midrule
ISIC-2017 & 0.30 & 0.0 & flip4 & $\checkmark$ & $\checkmark$ \\
ISIC-2018 & 0.25 & 0.0 & flip4 & $\checkmark$ & $\checkmark$ \\
PH2 & 0.06 & 0.0 & flip4 & $\checkmark$ & $\checkmark$ \\
\bottomrule
\end{tabular}
}
\end{table}

The test set is never used for operating-point selection. For the final RABC benchmark results, validation search selects $\sigma=0$ on all three datasets, which is consistent with the learned calibration branch absorbing the boundary correction needed by the deployed model without relying on additional Gaussian smoothing at test time. The no-RABC controls reported later are retained to quantify how much performance can still be obtained without any target-domain operating-point tuning.

The evaluation protocol enforces the following constraints:

\begin{itemize}[leftmargin=*]
\item \textbf{Annotation-free training}: no manual pixel-level annotations are used at any training stage; all supervision derives from automatically generated pseudo-labels.
\item \textbf{Strict target-label exclusion during adaptation}: for ISIC-2018 and PH2, ground-truth masks are never used for gradient-based training, pseudo-label generation, or Stage 2 interaction-branch adaptation.
\item \textbf{Validation-only model selection}: operating points are selected on validation labels only; the test set never participates in tuning.
\item \textbf{Comparison and statistics}: literature baselines are reported as benchmark-level margins under their published protocols; paired inference is restricted to within-study settings with per-case predictions. Table~\ref{tab:no-target-label-main} summarizes controls with no target-domain validation labels.
\end{itemize}

\subsection{Benchmark Results}

Tables~\ref{tab:sota-isic2018}--\ref{tab:sota-ph2} position the full system against 11 published unsupervised baselines under matched dataset splits. Because several prior checkpoints and per-case predictions are unavailable, these tables are benchmark-level positioning rather than unified re-evaluation. The stronger within-study evidence comes from the reproduced RPI-Net anchor, Base/Dil./RABC comparison, target-label-free controls, repeated-seed analysis, and boundary-calibration measurements.

Results on ISIC-2018 (test, 1,000 samples) are listed in Table~\ref{tab:sota-isic2018}.

\begin{table}[htbp]
\centering
\scriptsize
\setlength{\tabcolsep}{3pt}
\renewcommand{\arraystretch}{1.08}
\caption{Comparison with unsupervised methods on ISIC-2018.}\label{tab:sota-isic2018}
\resizebox{\singlecoltablewidth}{!}{%
\begin{tabular}{@{}llcccc@{}}
\toprule
Method & ACC & DICE & JAC & SEN & SPE \\
\midrule
Sp. Merging & 84.53 & 70.11 & 68.89 & 69.82 & 92.79 \\
DRC & 83.85 & 68.71 & 56.05 & 70.00 & 97.00 \\
Saliency-CCE & 85.51 & 72.85 & 61.91 & 77.44 & 94.34 \\
A2S-v2 & 86.22 & 75.01 & 65.94 & 72.13 & 97.35 \\
SpecWRSC & 81.20 & 68.97 & 58.41 & 69.39 & 86.34 \\
NCut & 82.48 & 69.11 & 58.78 & 68.12 & 88.31 \\
K-means & 83.70 & 71.46 & 61.57 & 71.73 & 88.09 \\
SGSCN & 82.30 & 70.97 & 61.81 & 71.17 & 87.36 \\
SLED & 86.93 & 77.68 & 69.35 & 80.12 & 91.84 \\
USL-Net & 88.45 & 78.79 & 68.32 & 90.91 & 87.76 \\
RPI-Net & 90.21 & 82.88 & 74.14 & 87.17 & 94.05 \\
MedSAM (box) $\dagger$ & 85.94 & 65.26 & 53.21 & 53.88 & 98.91 \\
\textbf{Ours} & \textbf{91.93} & \textbf{85.67} & \textbf{79.14} & 83.67 & \textbf{97.85} \\
\bottomrule
\end{tabular}
}
\end{table}

Results on ISIC-2017 (test, 600 samples) are listed in Table~\ref{tab:sota-isic2017}.

\begin{table}[htbp]
\centering
\scriptsize
\setlength{\tabcolsep}{3pt}
\renewcommand{\arraystretch}{1.08}
\caption{Comparison with unsupervised methods on ISIC-2017.}\label{tab:sota-isic2017}
\resizebox{\singlecoltablewidth}{!}{%
\begin{tabular}{@{}llcccc@{}}
\toprule
Method & ACC & DICE & JAC & SEN & SPE \\
\midrule
Sp. Merging & 79.89 & 54.66 & 46.02 & 59.16 & 88.72 \\
DRC & 83.77 & 59.11 & 45.40 & 70.35 & 95.58 \\
Saliency-CCE & 83.87 & 61.77 & 49.53 & 74.07 & 92.98 \\
A2S-v2 & 82.85 & 61.35 & 51.10 & 68.65 & 92.80 \\
SpecWRSC & 83.03 & 61.01 & 50.52 & 68.21 & 89.96 \\
NCut & 83.79 & 62.13 & 51.84 & 67.12 & 90.90 \\
K-means & 84.90 & 67.78 & 58.11 & 70.77 & 90.64 \\
SGSCN & 85.13 & 59.95 & 50.04 & 54.98 & 95.56 \\
SLED & 88.81 & 73.90 & 64.45 & 77.96 & 94.61 \\
USL-Net & 90.47 & 80.45 & 68.45 & 88.59 & 93.68 \\
RPI-Net & 87.62 & 75.84 & 66.21 & 84.15 & 91.73 \\
MedSAM (box) $\dagger$ & 87.67 & 65.96 & 53.35 & 53.98 & 99.49 \\
\textbf{Ours} & \textbf{92.08} & \textbf{81.90} & \textbf{72.86} & 81.57 & \textbf{97.21} \\
\bottomrule
\end{tabular}
}
\end{table}

Results on PH2 (test, 30 samples) are listed in Table~\ref{tab:sota-ph2}.

\begin{table}[htbp]
\centering
\scriptsize
\setlength{\tabcolsep}{3pt}
\renewcommand{\arraystretch}{1.08}
\caption{Comparison with unsupervised methods on PH2.}\label{tab:sota-ph2}
\resizebox{\singlecoltablewidth}{!}{%
\begin{tabular}{@{}llcccc@{}}
\toprule
Method & ACC & DICE & JAC & SEN & SPE \\
\midrule
Sp. Merging & 89.10 & 82.83 & 73.97 & 79.63 & 96.00 \\
DRC & 82.63 & 72.41 & 59.76 & 69.18 & 97.65 \\
Saliency-CCE & 84.62 & 76.79 & 65.62 & 78.90 & 93.77 \\
A2S-v2 & 85.66 & 78.21 & 67.15 & 87.56 & 91.37 \\
SpecWRSC & 85.96 & 78.12 & 68.54 & 79.08 & 89.74 \\
NCut & 88.15 & 80.17 & 70.51 & 76.16 & 94.94 \\
K-means & 91.64 & 86.14 & 77.78 & 83.82 & 95.03 \\
SGSCN & 91.80 & 87.89 & 80.15 & 84.26 & 93.82 \\
SLED & 93.00 & 90.34 & 83.50 & 89.97 & 96.37 \\
USL-Net & 92.43 & 88.90 & 80.11 & 93.62 & 93.07 \\
RPI-Net & 93.25 & 88.91 & 81.98 & 93.66 & 94.60 \\
MedSAM (box) $\dagger$ & 88.93 & 78.42 & 68.66 & 69.50 & 98.56 \\
\textbf{Ours} & \textbf{94.97} & \textbf{92.17} & \textbf{86.41} & \textbf{94.70} & 94.07 \\
\bottomrule
\end{tabular}
}
\end{table}

Across the three benchmark datasets, three patterns are observed:

\begin{itemize}[leftmargin=*]
\item The full system shows competitive matched-protocol positioning, with dataset-dependent SEN/SPE trade-offs.
\item Relative to RPI-Net \citep{ref1}, DICE increases by +2.79/+6.06/+3.26 pp on ISIC-2018/ISIC-2017/PH2 while reducing parameters from 105.12 M to 31.47 M.
\item MedSAM is included only as a prompt-based reference outside the annotation-free setting.
\end{itemize}

For deployment, the system keeps image-only inference, updates only 3.50\% of parameters during Stage 2, runs at 87.4 FPS, and selects $\sigma=0$ for all final operating points.

$\dagger$ MedSAM uses ground-truth-derived box prompts and is shown only as a prompt-based reference, not as an annotation-free baseline.

To complement the literature-level tables, we re-evaluated a reproduced RPI-Net checkpoint on ISIC-2018 under the same validation-selected search space. Table~\ref{tab:rpinet-anchor-isic2018} provides a controlled same-pipeline anchor for a strong uncertainty-aware baseline.

\begin{table}[htbp]
\centering
\scriptsize
\setlength{\tabcolsep}{3pt}
\renewcommand{\arraystretch}{1.08}
\caption{Same-pipeline anchor comparison on ISIC-2018 under the shared validation-selected inference search space.}\label{tab:rpinet-anchor-isic2018}
\resizebox{\singlecoltablewidth}{!}{%
\begin{tabular}{@{}lccccc@{}}
\toprule
Method & ACC & DICE & JAC & SEN & SPE \\
\midrule
Reproduced RPI-Net (same-pipeline) & 88.75 & 77.36 & 71.10 & 77.36 & 92.72 \\
\textbf{Ours} & \textbf{91.93} & \textbf{85.67} & \textbf{79.14} & \textbf{83.67} & \textbf{97.85} \\
\bottomrule
\end{tabular}
}
\par\vspace{2pt}
\begin{minipage}{\singlecoltablewidth}
\raggedright\footnotesize \textit{Note.} The reproduced RPI-Net anchor selected $\tau$=0.27 with TTA=\texttt{flip4}, keep-largest enabled, and no fill-holes under the shared validation search space. The final RABC model selected $\tau$=0.25 with TTA=\texttt{flip4}, keep-largest and fill-holes enabled. Relative to this same-pipeline anchor, the final model improves ACC/DICE/JAC/SEN/SPE by +3.18/+8.31/+8.04/+6.31/+5.13 percentage points, respectively.
\end{minipage}
\end{table}

\begin{figure}[htbp]
\centering
\includegraphics[width=\linewidth]{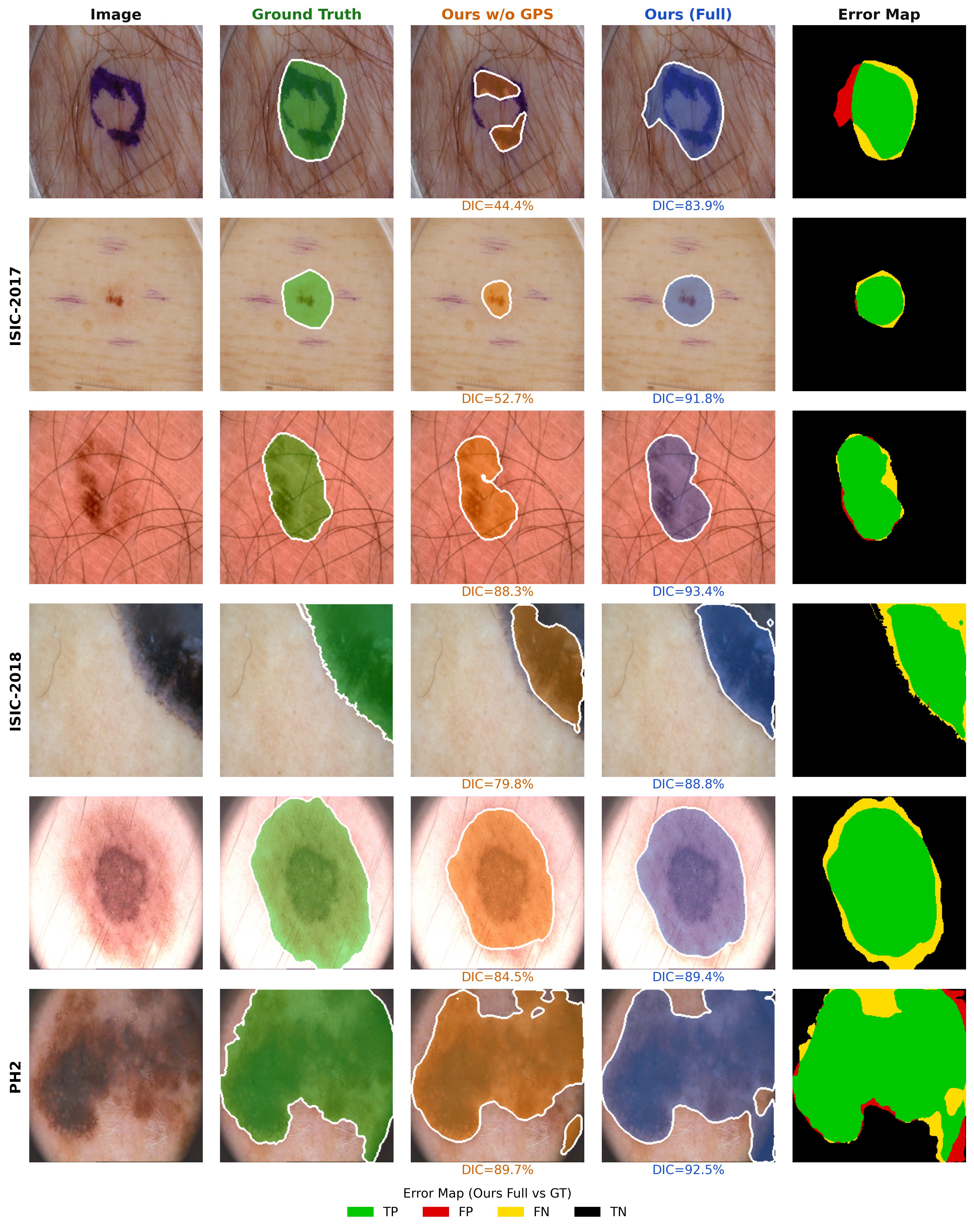}
\caption{Qualitative segmentation comparison. Columns show image, GT, raw P0, final prediction, and error map (green=TP, red=FP, yellow=FN).}\label{fig:qualitative-comparison}
\end{figure}

\begin{figure*}[t]
\centering
\includegraphics[width=0.98\textwidth]{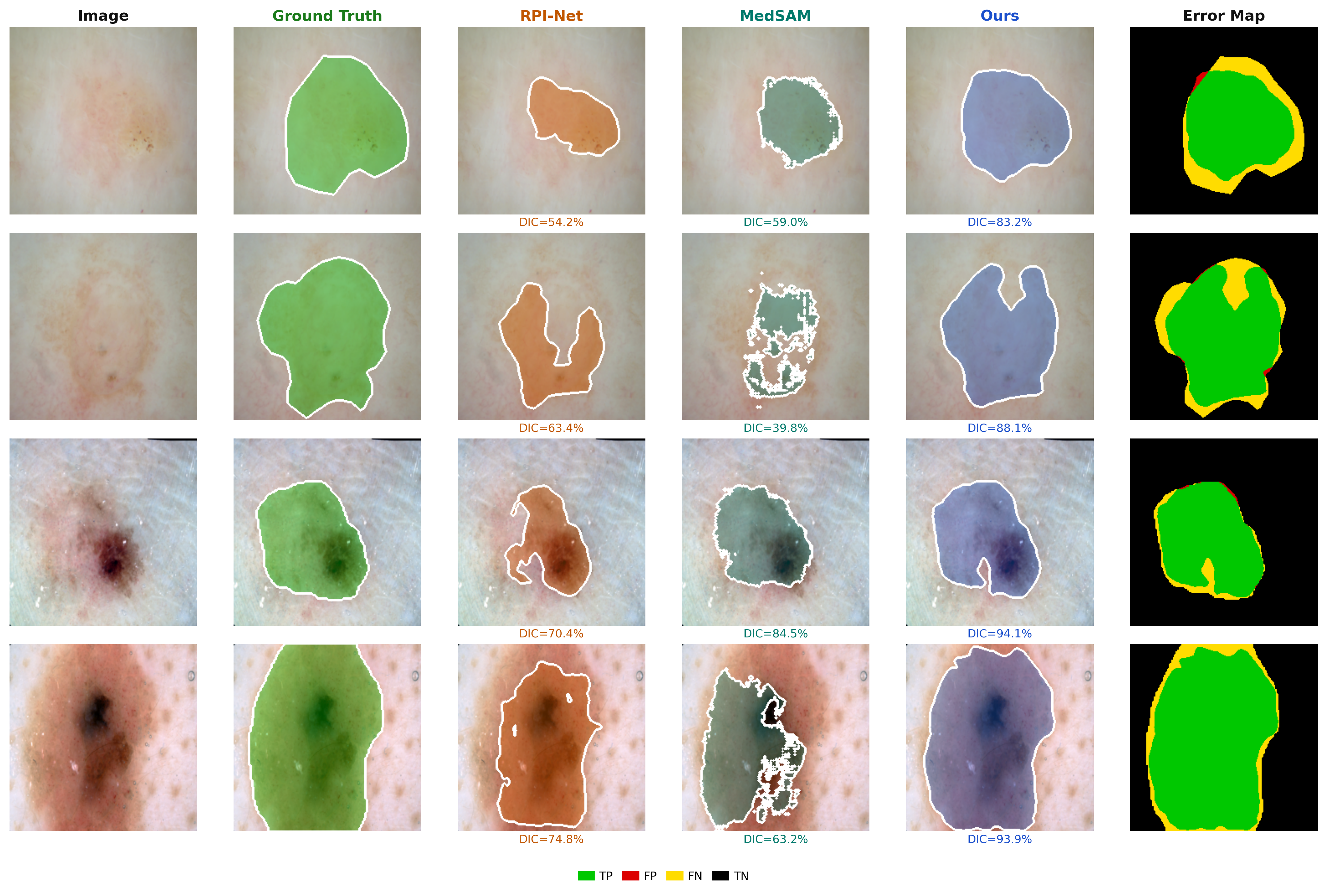}
\caption{Qualitative comparison on ISIC-2018. Columns show image, GT, reproduced RPI-Net anchor, prompted MedSAM reference, final prediction, and final-model error map.}
\label{fig:qualitative-prior-baseline}
\end{figure*}

Qualitative examples in Figures~\ref{fig:qualitative-comparison} and \ref{fig:qualitative-prior-baseline} illustrate boundary refinement relative to raw P0 and the reproduced RPI-Net anchor.

\subsection{Ablation Studies}

We first examine the training-side strategy on ISIC-2018 (Table~\ref{tab:ablation-training-isic2018}), with annotation-free training throughout and validation-based threshold selection by JAC.

\begin{table*}[t]
\centering
\scriptsize
\setlength{\tabcolsep}{2pt}
\renewcommand{\arraystretch}{1.08}
\caption{Training-side ablation (A0--A2) on ISIC-2018. A2 includes Stage 2 interaction-branch adaptation together with its own validation-selected operating point.}\label{tab:ablation-training-isic2018}
\begin{tabularx}{\textwidth}{@{}Z{0.06\textwidth}YZ{0.07\textwidth}Z{0.09\textwidth}Z{0.09\textwidth}Z{0.09\textwidth}Z{0.09\textwidth}Z{0.09\textwidth}@{}}
\toprule
ID & Configuration & Thr & ACC & DICE & JAC & SEN & SPE \\
\midrule
A0 & Baseline (ConvNeXt-Tiny + IPC/PIA + DUL) & 0.65 & 87.76 & 76.37 & 67.57 & 76.04 & 94.26 \\
A1 & A0 + Source distillation & 0.50 & 90.30 & 82.15 & 74.20 & 79.66 & 97.01 \\
A2 & A1 + Stage 2 interaction-branch adaptation & 0.37 & \textbf{91.40} & \textbf{85.08} & \textbf{77.83} & \textbf{83.01} & \textbf{98.02} \\
\bottomrule
\end{tabularx}
\end{table*}

Key observations:

\begin{itemize}[leftmargin=*]
\item Source distillation (A0$\rightarrow$A1) gives the largest single gain: $\Delta$DICE=+5.78 pp and $\Delta$JAC=+6.63 pp.
\item The full Stage 2 protocol (A1$\rightarrow$A2) adds a further $\Delta$DICE=+2.93 pp while increasing SPE from 97.01\% to 98.02\%; this pattern is more consistent with a protocol-level gain than with a parameter-update-only effect.
\item Overall, the two-stage design yields cumulative gains of $\Delta$DICE=+8.71 pp and $\Delta$JAC=+10.26 pp.
\end{itemize}

To isolate training effects from validation-selected threshold changes, we additionally re-evaluated A0/A1/A2 under a shared fixed operating point ($\tau$=0.30, TTA=\texttt{none}, no morphology). The resulting DICE/JAC values increase monotonically from 75.19/65.63 (A0) to 81.98/73.82 (A1) and 85.16/77.92 (A2), while SEN/SPE improve from 83.07/88.84 to 83.67/94.59 and 84.04/97.56, respectively. This fixed-operating-point check confirms that the training-side gain is not merely an artifact of per-model threshold selection.

Paired bootstrap resampling confirms that the cumulative A0$\rightarrow$A2 gain is statistically significant (p < 0.0002).

\begin{figure}[htbp]
\centering
\includegraphics[width=\linewidth]{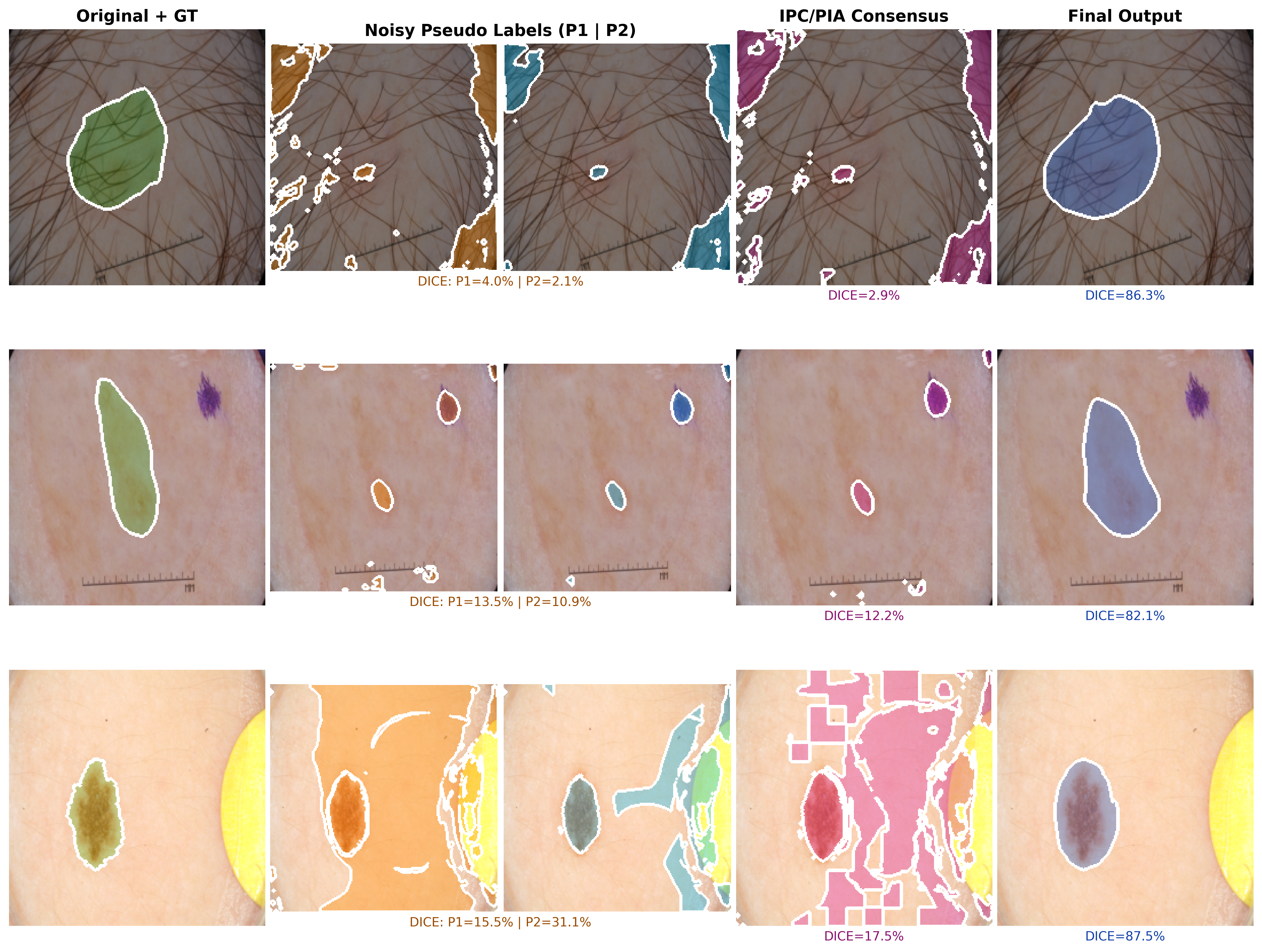}
\caption{Pseudo-label refinement example. IPC/PIA converts divergent pseudo-labels into a reliability-weighted consensus before final prediction.}
\label{fig:pseudo-flow}
\end{figure}

We next quantify the contribution of each lightweight inference component on the no-RABC A2 backbone by incrementally applying post-processing steps from the raw output (P0) to the reference pipeline (P4), as summarized in Tables~\ref{tab:ablation-infer-isic2017}--\ref{tab:ablation-infer-ph2}. These tables show what thresholding, morphology, optional reference smoothing, and TTA can achieve before adding RABC. The "15/15" column marks improvement over reproduced RPI-Net on all 15 overlap metrics across the three datasets.

For ISIC-2017 (test, 600 samples), the staged inference results are listed in Table~\ref{tab:ablation-infer-isic2017}.

\begin{table*}[t]
\centering
\scriptsize
\setlength{\tabcolsep}{2pt}
\renewcommand{\arraystretch}{1.08}
\caption{Inference-side ablation pipeline on ISIC-2017, including optional reference smoothing.}\label{tab:ablation-infer-isic2017}
\begin{tabularx}{\textwidth}{@{}Z{0.05\textwidth}YZ{0.06\textwidth}Z{0.06\textwidth}Z{0.07\textwidth}Z{0.065\textwidth}Z{0.065\textwidth}Z{0.065\textwidth}Z{0.065\textwidth}Z{0.065\textwidth}Z{0.07\textwidth}@{}}
\toprule
ID & Configuration & $\tau$ & $\sigma$ & TTA & ACC & DICE & JAC & SEN & SPE & 15/15 \\
\midrule
P0 & Raw & 0.30 & 0.0 & none & 90.88 & 78.81 & 68.44 & 73.38 & 98.32 & $\times$ \\
P1 & + Threshold tuning & 0.02 & 0.0 & none & 90.61 & 77.64 & 67.12 & 87.43 & 92.77 & $\times$ \\
P2 & + fill-holes + keep-largest & 0.02 & 0.0 & none & 91.79 & 80.73 & 71.88 & 84.74 & 94.52 & $\checkmark$ \\
P3 & + reference smoothing ($\sigma$=1.0) & 0.02 & 1.0 & none & 91.79 & 80.78 & 71.73 & 85.69 & 94.36 & $\checkmark$ \\
P4 & + TTA flip4 & 0.02 & 1.0 & flip4 & \textbf{91.83} & \textbf{81.03} & \textbf{72.05} & \textbf{86.45} & \textbf{94.08} & $\checkmark$ \\
\bottomrule
\end{tabularx}
\end{table*}

For ISIC-2018 (test, 1000 samples), the staged inference results are listed in Table~\ref{tab:ablation-infer-isic2018}.

\begin{table*}[t]
\centering
\scriptsize
\setlength{\tabcolsep}{2pt}
\renewcommand{\arraystretch}{1.08}
\caption{Inference-side ablation pipeline on ISIC-2018, including optional reference smoothing.}\label{tab:ablation-infer-isic2018}
\begin{tabularx}{\textwidth}{@{}Z{0.05\textwidth}YZ{0.06\textwidth}Z{0.06\textwidth}Z{0.07\textwidth}Z{0.065\textwidth}Z{0.065\textwidth}Z{0.065\textwidth}Z{0.065\textwidth}Z{0.065\textwidth}Z{0.07\textwidth}@{}}
\toprule
ID & Configuration & $\tau$ & $\sigma$ & TTA & ACC & DICE & JAC & SEN & SPE & 15/15 \\
\midrule
P0 & Raw & 0.30 & 0.0 & none & 91.47 & 85.16 & 77.91 & 84.04 & 97.56 & $\times$ \\
P1 & + Threshold tuning & 0.13 & 0.0 & none & \textbf{91.13} & \textbf{84.22} & \textbf{76.50} & \textbf{87.28} & \textbf{94.96} & $\checkmark$ \\
\bottomrule
\end{tabularx}
\end{table*}

As shown in Table~\ref{tab:ablation-infer-isic2018}, ISIC-2018 requires only threshold tuning to improve on all five RPI-Net metrics; within this operating-point study, optional reference smoothing and TTA are unnecessary for this dataset. The operating point is selected by validation JAC under the unified protocol rather than by maximizing DICE alone, which explains why P1 slightly lowers DICE relative to raw P0 while still improving SEN under the selected protocol.

For PH2 (test, 30 samples), the staged inference results are listed in Table~\ref{tab:ablation-infer-ph2}.

\begin{table*}[t]
\centering
\scriptsize
\setlength{\tabcolsep}{2pt}
\renewcommand{\arraystretch}{1.08}
\caption{Inference-side ablation pipeline on PH2, including optional reference smoothing.}\label{tab:ablation-infer-ph2}
\begin{tabularx}{\textwidth}{@{}Z{0.05\textwidth}YZ{0.06\textwidth}Z{0.06\textwidth}Z{0.07\textwidth}Z{0.065\textwidth}Z{0.065\textwidth}Z{0.065\textwidth}Z{0.065\textwidth}Z{0.065\textwidth}Z{0.07\textwidth}@{}}
\toprule
ID & Configuration & $\tau$ & $\sigma$ & TTA & ACC & DICE & JAC & SEN & SPE & 15/15 \\
\midrule
P0 & Raw & 0.30 & 0.0 & none & 94.18 & 91.63 & 85.46 & 89.70 & 98.23 & $\times$ \\
P1 & + Threshold tuning & 0.10 & 0.0 & none & 92.98 & 86.94 & 77.88 & 93.17 & 92.66 & $\times$ \\
P2 & + fill-holes + keep-largest & 0.10 & 0.0 & none & 94.71 & 92.15 & 86.38 & 93.03 & 95.55 & $\times$ \\
P3 & + reference smoothing ($\sigma$=1.0) & 0.10 & 1.0 & none & 94.69 & 91.98 & 86.09 & 93.57 & 95.26 & $\times$ \\
P4 & + TTA flip4 & 0.10 & 1.0 & flip4 & \textbf{94.77} & \textbf{92.14} & \textbf{86.35} & \textbf{93.78} & \textbf{95.09} & $\checkmark$ \\
\bottomrule
\end{tabularx}
\end{table*}

\begin{figure*}[!t]
\centering
\includegraphics[width=0.98\textwidth]{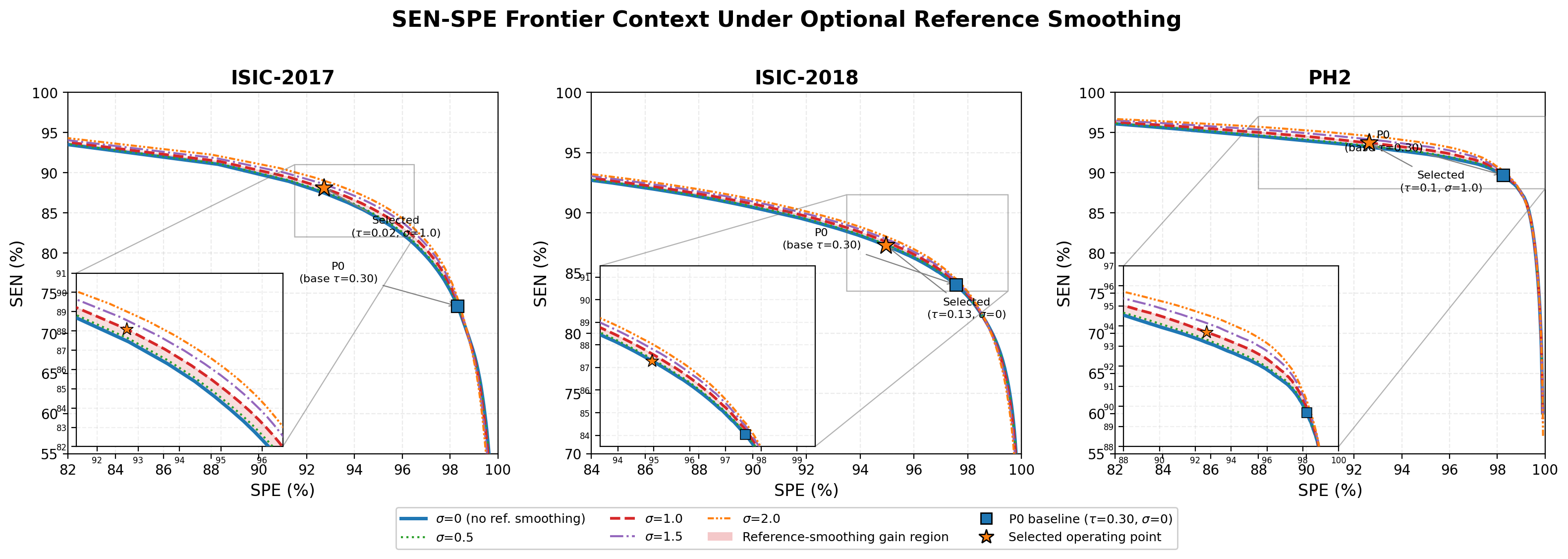}
\caption{SEN--SPE frontiers for optional reference smoothing on the no-RABC backbone. RABC is evaluated separately as learned calibration.}\label{fig:gps-frontier}
\end{figure*}

Across datasets, the P0$\rightarrow$P4 reference pipeline shows the expected recall-specificity trade-off: thresholding raises SEN, morphology recovers precision, optional smoothing acts as a nonlearned comparator, and TTA improves stability. Figure~\ref{fig:gps-frontier} summarizes this fixed post-processing behavior on the no-RABC backbone.

To test whether RABC can be reduced to a crude binary-mask expansion rule, we compare three variants under their own validation-selected operating points: the selected no-RABC baseline, the selected binary-dilation baseline, and RABC. Unlike the literature baselines in Tables~\ref{tab:sota-isic2018}--\ref{tab:sota-ph2}, this comparison is fully controlled within the present study and therefore isolates the effect of boundary calibration under a common evaluation pipeline. Table~\ref{tab:rabc-vs-dilation} shows that validation search never selects dilation as the preferred operating point, the selected dilation variant remains below the selected no-RABC baseline on DICE/JAC across all three datasets, and RABC yields more favorable results than dilation on all three datasets. Relative to the selected no-RABC baseline, however, the RABC gain is modest and dataset-dependent, which is why we interpret it as a controlled calibration benefit rather than as the sole source of the benchmark improvement.

On ISIC-2018, a paired bootstrap comparison between the selected dilation baseline and RABC further shows significant gains in JAC and SPE for RABC (two-sided $p=0.011$ and $p=0.017$, respectively), while the DICE gain remains borderline ($p=0.060$). This supports the view that RABC provides a cleaner calibration trade-off than binary dilation on the most stable external dataset, even when the absolute overlap margin is modest.

Because Table~\ref{tab:rabc-vs-dilation} still allows each method its own validation-selected operating point, we ran one stricter control in which the no-RABC protocol was fixed first and only one factor was changed at a time. On ISIC-2018, we fixed the selected no-RABC protocol ($\tau=0.21$, TTA=\texttt{flip4}, \texttt{keep-largest}=true, \texttt{fill-holes}=false), then matched the validation SEN of RABC by either lowering the no-RABC threshold globally or adding a single dilation step. Table~\ref{tab:rabc-fixed-protocol} shows that both crude alternatives lose JAC and SPE relative to RABC under this fixed protocol. On ISIC-2017, the nearest SEN-matched global threshold remained the same as the base threshold ($\tau=0.28$), while RABC still produced the most favorable overlap under the shared protocol. This behavior is consistent with RABC acting as a learned local calibration mechanism rather than a generic global recall push.

\begin{table}[htbp]
\centering
\scriptsize
\setlength{\tabcolsep}{3pt}
\renewcommand{\arraystretch}{1.06}
\caption{Fixed-protocol ISIC-2018 control. The selected no-RABC protocol is fixed first, then threshold shift and dilation are matched to the validation SEN of RABC instead of being tuned for overlap.}\label{tab:rabc-fixed-protocol}
\resizebox{\singlecoltablewidth}{!}{%
\begin{tabular}{@{}lcccc@{}}
\toprule
Method & DICE & JAC & SEN & SPE \\
\midrule
Base & 85.66 & 79.14 & 84.20 & \textbf{97.49} \\
Threshold shift ($\tau=0.13$) & 84.76 & 78.31 & 84.83 & 96.18 \\
Dilation ($1$ iter) & 85.30 & 78.50 & \textbf{85.42} & 96.87 \\
RABC & \textbf{85.67} & \textbf{79.16} & 84.25 & 97.46 \\
\bottomrule
\end{tabular}
}
\end{table}

\vspace{-2pt}
\begin{table*}[t]
\centering
\scriptsize
\setlength{\tabcolsep}{2pt}
\renewcommand{\arraystretch}{1.04}
\caption{Controlled within-study comparison of Base, validation-selected binary dilation (Dil.), and RABC under the same evaluation pipeline. Lower is better for HD95/ASSD.}\label{tab:rabc-vs-dilation}
\begin{tabularx}{\textwidth}{@{}Z{0.11\textwidth}Z{0.17\textwidth}Z{0.07\textwidth}Z{0.07\textwidth}Z{0.07\textwidth}Z{0.07\textwidth}Z{0.09\textwidth}Z{0.09\textwidth}@{}}
\toprule
Dataset & Method & DICE & JAC & SEN & SPE & HD95 $\downarrow$ & ASSD $\downarrow$ \\
\midrule
ISIC-2017 & Base & 81.83 & 72.75 & 81.34 & 97.27 & 20.49 & 9.13 \\
ISIC-2017 & Dil. & 81.67 & 72.55 & \textbf{81.68} & \textbf{97.35} & \textbf{20.44} & 9.16 \\
ISIC-2017 & RABC & \textbf{81.90} & \textbf{72.86} & 81.57 & 97.21 & \textbf{20.44} & \textbf{9.09} \\
\midrule
ISIC-2018 & Base & 85.66 & 79.14 & 84.20 & 97.49 & 21.29 & 8.96 \\
ISIC-2018 & Dil. & 85.01 & 78.16 & \textbf{84.83} & 97.08 & 21.42 & 9.30 \\
ISIC-2018 & RABC & \textbf{85.68} & \textbf{79.15} & 83.68 & \textbf{97.85} & \textbf{20.93} & \textbf{8.84} \\
\midrule
PH2 & Base & \textbf{92.28} & \textbf{86.59} & 93.90 & 94.96 & 11.79 & 4.58 \\
PH2 & Dil. & 92.01 & 86.14 & 94.10 & \textbf{95.03} & 12.28 & 4.71 \\
PH2 & RABC & 92.17 & 86.41 & \textbf{94.70} & 94.07 & \textbf{11.31} & \textbf{4.53} \\
\bottomrule
\end{tabularx}
\end{table*}
\vspace{-3pt}

We also perform a leave-one-out decoder ablation on ISIC-2017 (Table~\ref{tab:ablation-decoder-isic2017}). Each variant is trained from scratch and evaluated at the fixed raw operating point ($\tau$=0.30, no post-processing).

\begin{table}[htbp]
\centering
\scriptsize
\setlength{\tabcolsep}{3pt}
\renewcommand{\arraystretch}{1.08}
\caption{Decoder module-group ablation on ISIC-2017.}\label{tab:ablation-decoder-isic2017}
\resizebox{\singlecoltablewidth}{!}{%
\begin{tabular}{@{}llcccc@{}}
\toprule
Configuration & ACC & DICE & JAC & SEN & SPE \\
\midrule
Full (all modules enabled) & \textbf{90.88} & \textbf{78.81} & \textbf{68.44} & 73.38 & \textbf{98.32} \\
- Context Block (CSCB) & 86.64 & 68.32 & 57.59 & 78.17 & 90.80 \\
- Detail Block (DAB) & 87.08 & 69.06 & 58.07 & 75.92 & 91.46 \\
- Boundary Block (BRB) & 86.78 & 68.69 & 57.97 & 77.13 & 90.98 \\
- All Decoder Modules & 87.03 & 70.11 & 59.14 & 80.14 & 89.56 \\
\bottomrule
\end{tabular}
}
\end{table}

Key observations:

\begin{enumerate}[leftmargin=*]
\item The full decoder yields higher overlap than every single-removal variant, confirming that the three module groups act cooperatively rather than redundantly.
\item Removing CSCB causes the largest overlap drop, indicating that suppressing background-driven responses is the largest single contribution.
\item Across all removals, SEN rises while SPE falls, showing that the decoder primarily improves precision and boundary discipline rather than raw sensitivity.
\end{enumerate}

\begin{figure}[htbp]
\centering
\includegraphics[width=\linewidth]{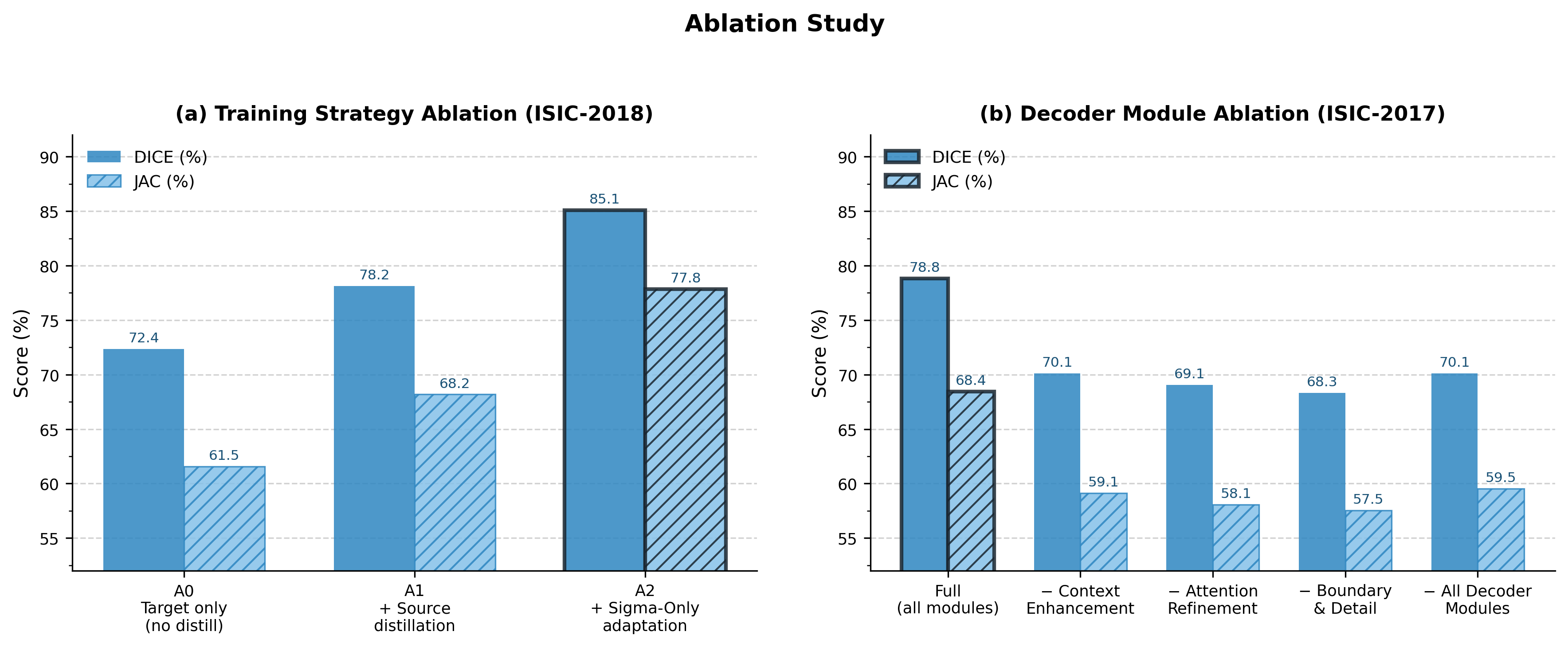}
\caption{Ablation visualization for training strategy and decoder module groups.}\label{fig:ablation-visualization}
\end{figure}

Figure~\ref{fig:ablation-visualization} summarizes the training and decoder ablations.

\subsection{Additional Analyses}

To examine the role of target-domain validation labels, Table~\ref{tab:no-target-label-main} reports controls that avoid target-domain validation tuning for both the no-RABC backbone and the final RABC system.

\begin{table}[htbp]
\centering
\scriptsize
\setlength{\tabcolsep}{3pt}
\renewcommand{\arraystretch}{1.08}
\caption{Target-domain controls without target-domain validation labels. Raw-P0 uses $\tau=0.30$ without TTA or morphology. Source-transferred protocols select one operating point on ISIC-2017 validation and reuse it on both targets.}\label{tab:no-target-label-main}
\resizebox{\singlecoltablewidth}{!}{%
\begin{tabular}{@{}lllcccc@{}}
\toprule
Target dataset & Method & Protocol & DICE & JAC & SEN & SPE \\
\midrule
ISIC-2018 & no-RABC & Raw-P0 fixed & 85.16 & 77.91 & 84.04 & 97.56 \\
ISIC-2018 & no-RABC & Source-transferred shared & 83.97 & 77.41 & 85.70 & 94.52 \\
ISIC-2018 & RABC & Raw-P0 fixed & 85.17 & 77.92 & 84.09 & 97.53 \\
ISIC-2018 & RABC & Source-transferred shared & \textbf{85.51} & \textbf{78.94} & 82.90 & \textbf{98.23} \\
\midrule
PH2 & no-RABC & Raw-P0 fixed & 91.63 & 85.46 & 89.70 & \textbf{98.23} \\
PH2 & no-RABC & Source-transferred shared & \textbf{92.28} & \textbf{86.59} & \textbf{93.90} & 94.96 \\
PH2 & RABC & Raw-P0 fixed & 91.63 & 85.46 & 89.71 & 98.22 \\
PH2 & RABC & Source-transferred shared & 91.68 & 85.60 & 89.57 & 98.03 \\
\bottomrule
\end{tabular}
}
\end{table}

These controls define the target-label-free boundary. On ISIC-2018, source-transferred RABC improves DICE/JAC over no-RABC by 1.54/1.53 points and raises SPE from 94.52\% to 98.23\%. On PH2, it remains below the selected no-RABC source-transferred control, indicating dataset-dependent utility.

\begin{table}[htbp]
\centering
\scriptsize
\setlength{\tabcolsep}{3pt}
\renewcommand{\arraystretch}{1.08}
\caption{Target-free triplet comparison using operating points selected only on ISIC-2017 validation data.}\label{tab:targetfree-triplet}
\resizebox{\singlecoltablewidth}{!}{%
\begin{tabular}{@{}llcccc@{}}
\toprule
Target dataset & Method & DICE & JAC & SEN & SPE \\
\midrule
ISIC-2018 & no-RABC & 83.97 & 77.41 & 85.71 & 94.52 \\
ISIC-2018 & Dil. & 83.25 & 76.18 & \textbf{86.50} & 93.66 \\
ISIC-2018 & RABC & \textbf{85.51} & \textbf{78.94} & 82.90 & \textbf{98.23} \\
\midrule
PH2 & no-RABC & \textbf{92.28} & \textbf{86.59} & 93.90 & 94.96 \\
PH2 & Dil. & 91.76 & 85.70 & \textbf{94.97} & 94.34 \\
PH2 & RABC & 91.68 & 85.60 & 89.57 & \textbf{98.03} \\
\bottomrule
\end{tabular}
}
\end{table}

Table~\ref{tab:targetfree-triplet} compares no-RABC, binary dilation, and RABC under fixed source-selected operating points. On ISIC-2018, RABC improves DICE/JAC over no-RABC by 1.54/1.53 points ($p=0.0012/0.0010$) and over dilation by 2.26/2.77 points ($p<10^{-4}$). On PH2, RABC does not improve overlap over no-RABC. This confirms target-free utility in one external setting and dataset dependence in the smaller PH2 split.

Repeating the PH2 target-free protocols across three resumed seeds preserved the ranking with negligible variation, suggesting operating-point mismatch on the small split rather than unstable optimization.

The final RABC results have macro-average ACC/DICE/JAC/SEN/SPE of 92.99/86.58/79.47/86.65/96.38 across the three datasets, with strongest margins on overlap metrics and lower macro SEN due to more specificity-preserving operating points.

Across three resumed ISIC-2018 seeds, no-RABC achieved mean ACC/DICE/JAC/SEN/SPE of 91.94/85.66/79.13/83.62/97.89, whereas RABC achieved 92.01/85.77/79.28/84.08/97.63. ACC and SEN gains were consistently significant, while DICE/JAC gains were modest, supporting RABC as a stable calibration mechanism.

\begin{table}[htbp]
\centering
\scriptsize
\setlength{\tabcolsep}{3pt}
\renewcommand{\arraystretch}{1.08}
\caption{Boundary-band probability calibration before thresholding or morphology; lower is better.}\label{tab:rabc-calibration}
\resizebox{\singlecoltablewidth}{!}{%
\begin{tabular}{@{}lcccccc@{}}
\toprule
Dataset & \multicolumn{3}{c}{no-RABC} & \multicolumn{3}{c}{RABC} \\
\cmidrule(lr){2-4}\cmidrule(lr){5-7}
 & ECE & Brier & NLL & ECE & Brier & NLL \\
\midrule
ISIC-2017 & 0.3135 & 0.3142 & 1.1256 & \textbf{0.3083} & \textbf{0.3095} & \textbf{1.1025} \\
ISIC-2018 & 0.2908 & 0.2847 & 1.0877 & \textbf{0.2902} & \textbf{0.2841} & \textbf{1.0829} \\
PH2 & 0.2953 & 0.2945 & 1.0225 & \textbf{0.2951} & \textbf{0.2943} & \textbf{1.0209} \\
\bottomrule
\end{tabular}
}
\end{table}

Table~\ref{tab:rabc-calibration} reports calibration-specific evidence in probability space on a narrow ground-truth boundary band. RABC lowers ECE, Brier score, and NLL on all three datasets, with the clearest effect on ISIC-2017 and near-neutral changes on PH2. This supports RABC as a localized calibration module with dataset-dependent utility.

\begin{figure*}[t]
\centering
\includegraphics[width=0.92\textwidth]{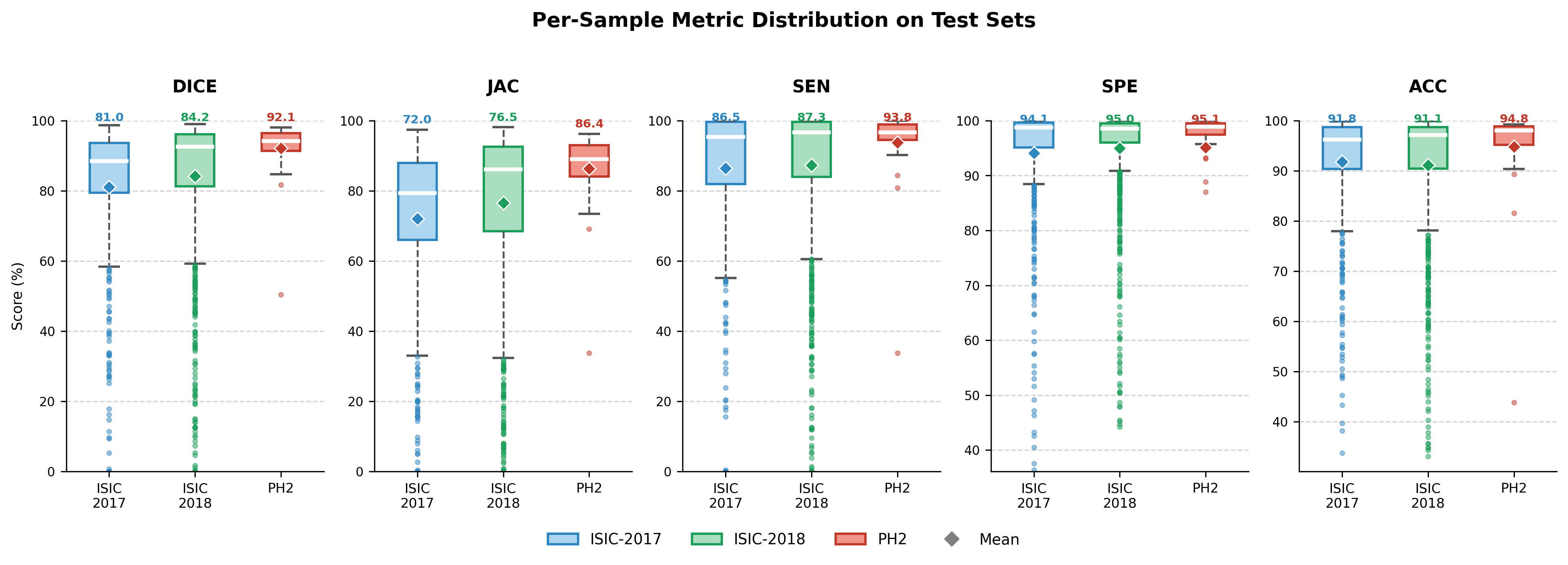}
\caption{Per-sample metric distributions on the three test sets.}\label{fig:metric-boxplot}
\end{figure*}

Figure~\ref{fig:metric-boxplot} shows per-sample metric distributions.

Table~\ref{tab:boundary-distance} reports boundary distance metrics at the final operating point against a reproduced RPI-Net baseline under each method's own validation-selected operating point:

\begin{table}[htbp]
\centering
\scriptsize
\setlength{\tabcolsep}{3pt}
\renewcommand{\arraystretch}{1.08}
\caption{Boundary distance comparison against a reproduced RPI-Net baseline.}\label{tab:boundary-distance}
\resizebox{\singlecoltablewidth}{!}{%
\begin{tabular}{@{}llcccc@{}}
\toprule
Dataset & Method & DICE & JAC & HD95 & ASSD \\
\midrule
ISIC-2017 & Reproduced RPI-Net & 68.83 & 60.06 & 38.02 & 19.26 \\
ISIC-2017 & Ours & \textbf{81.90} & \textbf{72.86} & \textbf{20.44} & \textbf{9.09} \\
ISIC-2018 & Reproduced RPI-Net & 77.38 & 71.14 & 34.33 & 15.88 \\
ISIC-2018 & Ours & \textbf{85.67} & \textbf{79.14} & \textbf{20.93} & \textbf{8.84} \\
PH2 & Reproduced RPI-Net & 85.77 & 80.02 & 22.30 & 10.45 \\
PH2 & Ours & \textbf{92.17} & \textbf{86.41} & \textbf{11.31} & \textbf{4.53} \\
\bottomrule
\end{tabular}
}
\end{table}

The reproduced baseline indicates a clear boundary-localization advantage on all three datasets, where both HD95 and ASSD decrease together with higher overlap metrics. This pattern is consistent with the intended role of RABC: the calibration branch does not simply inflate the foreground region, but improves contour placement under the shared evaluation protocol.

\begin{table}[htbp]
\centering
\scriptsize
\setlength{\tabcolsep}{3pt}
\renewcommand{\arraystretch}{1.08}
\caption{Model efficiency comparison between RPI-Net and the proposed method.}\label{tab:model-efficiency}
\begin{tabular}{@{}llc@{}}
\toprule
Metric & RPI-Net & Ours \\
\midrule
Parameters (M) & 105.12 & \textbf{31.47} \\
GPU Memory (MB) & 484.73 & 686 \\
\bottomrule
\end{tabular}
\end{table}

Table~\ref{tab:model-efficiency} shows that our model uses only 29.9\% of RPI-Net's parameter count, at the cost of a higher full-resolution attention memory footprint.

To contextualize the remaining ceiling, we compare with six supervised references on the same dataset splits: U-Net, U-Net++, Att-DeepLabv3+, TransFuse, FAT-Net, and MALUNet \citep{ref10,ref11,ref14,ref13,ref15,ref16}. Values in Tables~\ref{tab:supervised-isic2017}--\ref{tab:supervised-ph2} follow the published benchmark summary of \citet{ref1} and are treated as literature upper bounds rather than retrained baselines.

\begin{table}[htbp]
\centering
\scriptsize
\setlength{\tabcolsep}{6pt}
\renewcommand{\arraystretch}{1.08}
\caption{Gap between the proposed annotation-free training framework and the mean DICE of six supervised reference models.}\label{tab:gap-supervised}
\resizebox{\singlecoltablewidth}{!}{%
\begin{tabular}{@{}lccc@{}}
\toprule
Dataset & Supervised mean DICE & Ours (unsup.) & Gap \\
\midrule
ISIC-2017 & 84.59 & 81.90 & 2.69 pp \\
ISIC-2018 & 88.01 & 85.67 & 2.34 pp \\
\textbf{PH2} & \textbf{93.16} & \textbf{92.17} & \textbf{0.99 pp} \\
\bottomrule
\end{tabular}
}
\end{table}

Most notably, on PH2 the gap to the supervised mean is only 0.99 percentage points.

Detailed supervised comparisons on ISIC-2017 (test, 600 samples) are listed in Table~\ref{tab:supervised-isic2017}.

\begin{table}[htbp]
\centering
\scriptsize
\setlength{\tabcolsep}{3pt}
\renewcommand{\arraystretch}{1.08}
\caption{Comparison with supervised methods on ISIC-2017.}\label{tab:supervised-isic2017}
\resizebox{\singlecoltablewidth}{!}{%
\begin{tabular}{@{}llcccc@{}}
\toprule
Method & ACC & DICE & JAC & SEN & SPE \\
\midrule
U-Net & 93.87 & 85.43 & 77.31 & 83.02 & 97.61 \\
U-Net++ & 93.24 & 84.45 & 76.05 & 80.78 & 98.60 \\
Att-DeepLabv3+ & 93.51 & 84.82 & 76.24 & 80.33 & 98.56 \\
TransFuse & 93.77 & 85.26 & 76.91 & 82.62 & 98.08 \\
FAT-Net & 93.41 & 84.97 & 76.51 & 81.06 & 98.51 \\
MALUNet & 92.32 & 82.59 & 73.57 & 82.12 & 97.79 \\
\textbf{Ours (unsup.)} & 92.08 & 81.90 & 72.86 & 81.57 & 97.21 \\
\bottomrule
\end{tabular}
}
\end{table}

Detailed supervised comparisons on ISIC-2018 (test, 1,000 samples) are listed in Table~\ref{tab:supervised-isic2018}.

\begin{table}[htbp]
\centering
\scriptsize
\setlength{\tabcolsep}{3pt}
\renewcommand{\arraystretch}{1.08}
\caption{Comparison with supervised methods on ISIC-2018.}\label{tab:supervised-isic2018}
\resizebox{\singlecoltablewidth}{!}{%
\begin{tabular}{@{}llcccc@{}}
\toprule
Method & ACC & DICE & JAC & SEN & SPE \\
\midrule
U-Net & 93.33 & 88.18 & 80.40 & 94.99 & 92.45 \\
U-Net++ & 93.72 & 88.23 & 80.65 & 93.86 & 94.34 \\
Att-DeepLabv3+ & 93.42 & 88.20 & 80.50 & 94.56 & 93.18 \\
TransFuse & 93.85 & 88.46 & 80.80 & 94.94 & 93.05 \\
FAT-Net & 93.73 & 88.95 & 81.87 & 92.75 & 94.92 \\
MALUNet & 92.07 & 86.04 & 77.87 & 90.89 & 94.57 \\
\textbf{Ours (unsup.)} & 91.93 & 85.67 & 79.14 & 83.67 & 97.85 \\
\bottomrule
\end{tabular}
}
\end{table}

Detailed supervised comparisons on PH2 (test, 30 samples) are listed in Table~\ref{tab:supervised-ph2}.

\begin{table}[htbp]
\centering
\scriptsize
\setlength{\tabcolsep}{3pt}
\renewcommand{\arraystretch}{1.08}
\caption{Comparison with supervised methods on PH2.}\label{tab:supervised-ph2}
\resizebox{\singlecoltablewidth}{!}{%
\begin{tabular}{@{}llcccc@{}}
\toprule
Method & ACC & DICE & JAC & SEN & SPE \\
\midrule
U-Net & 96.66 & 94.14 & 89.32 & 94.76 & 97.37 \\
U-Net++ & 95.18 & 90.80 & 83.61 & 95.25 & 93.05 \\
Att-DeepLabv3+ & 95.84 & 93.29 & 88.03 & 95.02 & 96.70 \\
TransFuse & 96.69 & 93.76 & 88.73 & 97.39 & 96.05 \\
FAT-Net & 96.49 & 93.63 & 88.93 & 96.39 & 96.03 \\
MALUNet & 96.45 & 93.32 & 87.89 & 96.07 & 96.26 \\
\textbf{Ours (unsup.)} & 94.97 & 92.17 & 86.41 & 94.70 & 94.07 \\
\bottomrule
\end{tabular}
}
\end{table}

Key observations:

\begin{itemize}[leftmargin=*]
\item The supervised gap remains modest on all three datasets and is smallest on PH2, where the DICE gap is only 0.99 pp.
\item These results suggest that, in favorable transfer settings, the proposed annotation-free framework can approach supervised performance without target-domain gradient supervision.
\end{itemize}

A supplementary qualitative comparison against the supervised TransFuse upper-bound reference on the same ISIC-2018 cases is provided in Supplementary Figure~S1.

\section{Discussion}

The main contribution is the reliability-aware deployment design rather than an isolated module gain. The system separates prior generation, reliability learning, restricted adaptation, and boundary calibration: training-time components absorb noisy pseudo supervision and target-domain reliability shifts, while the deployed image-only path remains stable. RABC is the most distinctive component in this design because it locally calibrates difficult contours after a reliable representation has been formed.

RABC differs from global threshold reduction and binary dilation by conditioning correction on local reliability. It predicts candidate regions from boundary confidence, uncertainty, and foreground probability, then regularizes the logit update through boundary-consistency, far-background-preservation, and sparsity constraints.

The dilation comparison supports this interpretation. RABC is more favorable than dilation on all three datasets, but its gain is dataset-dependent: ISIC-2017 shows modest overlap improvement, ISIC-2018 favors specificity and boundary-distance quality, and PH2 shows a recall-oriented trade-off. RABC should therefore be read as an operating-point calibration module rather than a source of uniform metric gains.

Accordingly, the strongest evidence for RABC comes from within-study controls: the reproduced RPI-Net anchor, Base/Dil./RABC comparison, fixed-protocol control, repeated-seed tests, and boundary-band calibration measurements. The literature tables provide system-level positioning only.

The higher final DICE on PH2 (92.17\%) than ISIC-2017 (81.90\%) likely reflects clearer PH2 boundaries and less distracting background structure. Interaction-branch adaptation targets such transfer settings by freezing source-domain semantics while updating reliability heads and the compact calibration branch. Reference smoothing remains only a lightweight comparator for fixed post-processing behavior.

Several limitations remain. Difficult cases include severe hair occlusion, very low lesion-to-skin contrast, and gradual boundaries. The darkness/centroid/area priors are dermoscopy-specific, although the reliability learning, adaptation, and calibration stages could use alternative prior generators. Evaluation is limited to public dermoscopy benchmarks and pixel-level metrics, and statistical inference is strongest for within-study ablations. Final performance also depends on validation-stage operating-point search, although no-target-label controls suggest that this search refines rather than enables performance. The supervised-gap analysis suggests a practical trade-off: when rapid transfer and low labeling cost matter more than the last few supervised percentage points, the annotation-free workflow remains useful.

\section{Conclusion}

We present RABC-Net, a reliability-aware annotation-free system in which RABC provides lightweight local boundary calibration. Interaction-branch adaptation, feature-decoupled decoding, and reliability-aware training stabilize the broader system. Across ISIC-2017, ISIC-2018, and PH2, the framework achieves macro-average DICE/JAC of 86.58\%/79.47\% and improves on validation-selected binary-dilation alternatives in controlled comparisons. The results support annotation-free dermoscopic segmentation as a practical low-annotation option and suggest that separating reliability learning, restricted adaptation, and local calibration is useful for deployment under noisy supervision. RABC contributes localized, calibration-oriented gains whose magnitude remains dataset dependent.

\section*{Declaration of Competing Interest}

The authors declare that they have no known competing financial interests or personal relationships that could have appeared to influence the work reported in this paper.

\section*{CRediT authorship contribution statement}

\textbf{Yujie Yao:} Conceptualization, Methodology, Software, Formal analysis, Visualization, Writing - original draft. \textbf{Yuhaohang He:} Software, Validation, Investigation, Data curation. \textbf{Junjie Huang:} Software, Formal analysis, Visualization. \textbf{Zhou Liu:} Investigation, Data curation, Resources. \textbf{Jiangzhao Li:} Methodology, Validation, Writing - review \& editing. \textbf{Yan Qiao:} Investigation, Data curation. \textbf{Wen Xiao:} Investigation, Resources. \textbf{Yunsen Liang:} Validation, Visualization. \textbf{Xiaofan Li:} Conceptualization, Supervision, Project administration, Writing - review \& editing.

\end{document}